%% file: main.tex
\documentclass[conference]{IEEEtran}
\IEEEoverridecommandlockouts
\usepackage{cite}
\usepackage{hyperref}
\usepackage{amsmath,amssymb,amsfonts}
\usepackage{algorithmic}
\usepackage{graphicx}
\usepackage{textcomp}
\usepackage{booktabs}
\usepackage{xcolor}
\usepackage{subcaption}
\usepackage{xspace}
\usepackage{siunitx}

\newcommand{\OSL}{\texttt{Osmotic Learning}\xspace}
\newcommand{\OLshort}{OSM\mbox{-L}\xspace}
\usepackage{tikz}
\usepackage{enumitem}
\newcommand{\numbercircle}[1]{%
    \tikz[baseline=(char.base)]{%
        \node[shape=circle, draw, fill=black, text=white, inner sep=0.15mm] (char) {#1};
    }%
}
\begin{document}

\title{Osmotic Learning: A Self-Supervised Paradigm \\for Decentralized Contextual Data Representation}


\author{
    \IEEEauthorblockN{Mario Colosi\IEEEauthorrefmark{1},
                      Reza Farahani\IEEEauthorrefmark{2},
                      Maria Fazio\IEEEauthorrefmark{1},
                      Radu Prodan\IEEEauthorrefmark{3},
                      Massimo Villari\IEEEauthorrefmark{1}}

    \vspace{0.2cm}

    \IEEEauthorblockA{\IEEEauthorrefmark{1}MIFT Department, University of Messina, Italy}

    \IEEEauthorblockA{\IEEEauthorrefmark{2}Institute of Information Technology, University of Klagenfurt, Austria}

    \IEEEauthorblockA{\IEEEauthorrefmark{3}Department of Computer Science, University of Innsbruck, Austria}
}



\maketitle

\input{tex/0-abstract}


\input{tex/1-introduction}
\input{tex/2-background}
\input{tex/3-ol-paradigm}
\input{tex/4-architecture}

\input{tex/5-evaluation-setup}

\input{tex/6-experimental-results}
\input{tex/7-sota}

\input{tex/8-discussion}

\input{tex/9-conclusions}

\section*{Acknowledgment}
This work received partial funding from the Italian Ministry of University and Research (MUR) ``Research projects of National Interest (PRIN-PNRR)'' call through the project ``Cloud Continuum aimed at On-Demand Services in Smart Sustainable Environments'' (CUP: J53D23015080001- IC: P2022YNBHP), and "SEcurity and RIghts in the CyberSpace (SERICS)" project (PE00000014), under the MUR National Recovery and Resilience Plan funded by the European Union - NextGenerationEU (CUP: D43C22003050001), and  the Austrian Research Promotion Agency (FFG), project 909989 ``AIM AT Stiftungsprofessur für Edge AI''.


\bibliographystyle{./bibliography/IEEEtran}
\bibliography{./bibliography/IEEEabrv}

\end{document}

%% file: tex/0-abstract.tex
\begin{abstract}
Data within a specific context gains deeper significance beyond its isolated interpretation.  In distributed systems, interdependent data sources reveal hidden relationships and latent structures, representing valuable information for many applications. This paper introduces \OSL (\OLshort), a self-supervised distributed learning paradigm designed to uncover higher-level latent knowledge from distributed data. The core of \OLshort is \textit{osmosis}, a process that synthesizes dense and compact representation by extracting contextual information, eliminating the need for raw data exchange between distributed entities. \OLshort iteratively aligns local data representations, enabling information diffusion and convergence into a dynamic equilibrium that captures contextual patterns. During training, it also identifies correlated data groups, functioning as a decentralized clustering mechanism.
Experimental results confirm \OLshort's  convergence and representation capabilities on structured datasets, achieving over \num{0.99} accuracy in local information alignment while preserving contextual integrity.

\end{abstract}

%% file: tex/1-introduction.tex
\section{Introduction}

In recent years, the exponential growth of data generated by distributed systems has necessitated the development of advanced learning paradigms to extract meaningful insights from decentralized data sources. Conventional centralized learning methods often face challenges, including data privacy, high communication overhead, and limited scalability~\cite{10278413}. 
This highlights the need for distributed learning systems to effectively leverage distributed data while extracting and synthesizing insights from the globally observed system.

Representation learning has emerged as a key process in machine learning (ML), enabling neural network (NN) models to extract and identify meaningful features from raw data \cite{reprlearning,10.1007/978-3-030-37309-2_8,ye2024ptarlprototypebasedtabularrepresentation}. This method transforms high-dimensional data into lower-dimensional vector space that captures the underlying semantics, improving the performance of downstream tasks. However, applying representation learning often brings complexities, particularly with unlabeled and non-independent and identically distributed (non-IID) data~\cite{wang2023doeslearningdecentralizednoniid}.
These challenges are particularly evident in distributed systems, where data are heterogeneous, fragmented, and not easily shareable. 
Despite these obstacles, representation learning has found diverse applications, such as \textit{(i)} microservice monitoring systems 
to identify global patterns and detect anomalies across services~\cite{9912640}; \textit{(ii)} proactive systems leveraging latent correlations between distributed data sources to predict events~\cite{10191086}; and \textit{(iii)} healthcare to facilitate the analysis of fragmented data across institutions while preserving data privacy \cite{mengyan}.

Inspired by the biological process of \textit{osmosis}, this paper introduces \OSL~(\OLshort), a novel self-supervised distributed learning paradigm that
uncovers higher-level latent knowledge from distributed data. As osmosis enables the natural flow between two separate environments to achieve equilibrium, \OLshort iteratively processes and aligns local data representations, allowing information to diffuse across distributed sources and converge toward a dynamic equilibrium of shared latent knowledge. 
This alignment enables the clustering of data sources containing correlated data, uncovering hidden structures and relationships. 

When interdependent data from different sources are considered through interconnected relationships, they acquire a broader meaning beyond their isolated interpretation, resulting in enriched information. \OLshort captures it as shared latent patterns, enabling the emergence of unified knowledge without requiring the sharing of raw data.
For example, consider a monitoring scenario with two roads: a weather sensor on one road measures rain, while a meter on the other detects congestion. Rain data alone provides no direct insight into the traffic status, yet there may be a strong correlation, e.g., heavy rain on the first road might coincide with traffic buildup on the second.
By contextualizing these data points, the traffic status on the second road can be inferred using only rain measurements from the first road.  
Although such correlation can be identified manually in simple scenarios, \OLshort automates the discovery of latent relationships across distributed complex systems, providing an abstract representation of global patterns and ensuring privacy and efficiency. 

The primary contributions of this work are: \textit{(i)} a theoretical formalization of the \OLshort paradigm; \textit{(ii)} the development of a worker-master architecture, where distributed entities interact with a central coordinator for implementing the \OLshort pipeline; \textit{(iii)} the design of a dynamic clustering mechanism for distributed data, enabling the discovery of latent structures; \textit{(iv)} a comprehensive experimental evaluation demonstrating the effectiveness of \OLshort, highlighting its adaptability to heterogeneous and distributed systems.

The paper has nine sections. Section~\ref{sec:background} discusses the fundamental background of \OLshort  followed by the formalization of the \OLshort paradigm in Section~\ref{sec:osmotic}. Section~\ref{sec:architecture} details  \OLshort system architecture, while Section~\ref{sec:setup} outlines the evaluation setup, leading to the experimental results in Section~\ref{sec:experiment}. Section~\ref{sec:sota} reviews how \OLshort relates to existing paradigms, followed by a discussion of  \OLshort limitations and potential future directions in Section~\ref{sec:discussion}. Section~\ref{sec:conclusion} finally concludes the paper.

%% file: tex/2-background.tex
\section{Background}
\label{sec:background}
This section introduces the foundational concepts of \OLshort, with key terms summarized in Table~\ref{tab:glossary}, covering notations from this and subsequent sections.
\paragraph{Context} refers to the set of interconnected elements that provide the background against which specific information acquires precise meaning.
\OLshort defines context as a set of correlated datasets represented by a shared latent structure emerging from their relationships to capture connections and patterns, offering a unified representation of system dynamics. 
Therefore, explicitly extracting and representing this structure transforms implicit dependencies into actionable insights, enabling applications like predictive modeling, anomaly detection, and distributed system optimization.

\paragraph{Sub-context}
in complex environments, elements within a context often form smaller subsets based on stronger interdependencies or localized relationships. While sub-contexts belong to the same global context, they can exhibit distinct behaviors or dynamics shaped by the unique characteristics and correlations of their data.

\paragraph{Embedding}
transforms raw inputs into a structured representation optimized for model processing, effectively capturing intrinsic relationships and underlying semantics within the data~\cite{6472238}. An embedding, indeed, is a dense, compact representation of data in a vector space, projecting relevant information into a fixed-size numerical format while preserving semantic relationships and reducing complexity.

\paragraph{Agent}
is an independent computational entity associated with a specific context, responsible for processing local data. An agent autonomously extracts higher-level representations from its embedding data, without requiring direct access to the data of other entities. This scheme indeed enhances the informational value of their local data, transforming it into meaningful contextual representations that capture the shared characteristics of their context or sub-context.
\begin{table}[t!]
\caption{Notations for \OLshort.}
\centering
\begin{tabular}{lcl}
\toprule
\textbf{Term}          & \textbf{Symbol} & \textbf{Description}                                \\ \midrule
Context                & \( C \)         & Shared structure among agents.                     \\
Sub-context            & \( C_k \)       & Cluster of correlated agents.  
                \\
Embedding              & \( e \)         & Dense data representation.                         \\
Agent                  & \( a_i \)       & Compute entity processing local data.                        \\
Local data             & \( \mathcal{X}_i \) & Features processed by an agent.               \\
Context Embedding      & \( e_{ctx} \)   & Global synthesis of local embeddings.              \\
Alignment Loss         & \( L_{align} \) & Promotes similarity to \( e_{ctx} \).              \\
Preservation Loss      & \( L_{pres} \)  & Keeps local information.  
            \\
Osmotic Strategy       & \( \Omega \)    & Synthesize a global context \( e_{ctx} \). \\

\bottomrule
\end{tabular}
\label{tab:glossary}
\end{table}

%% file: tex/3-ol-paradigm.tex
\section{\OSL Paradigm}
\label{sec:osmotic}

\OLshort is a self-supervised distributed learning paradigm designed to autonomously process local data and extract higher-level knowledge about the context to which the data pertains. 
\OLshort aims to express the meaning of local data within the context rather than focusing on their direct representation.
Through an iterative alignment process between local representations, it captures shared latent patterns, synthesizing context-level knowledge while ensuring raw data remains confined to its source.

\subsection{Formal Definition}
Let \( A=\{a_0,a_1,\dots,a_{n-1}\} \) be a set of $n$ agents belonging to a specific context $C$, where each agent corresponds to an independent compute instance processing local data through a designated local NN model. An agent $a_i$ is defined by the pair \( (\mathcal{X}_i, f_i) \), where $\mathcal{X}_i$ represents a dataset composed of a vector of local feature data, and $f_i$ is the NN model that processes these features. Specifically, the dataset $\mathcal{X}_i$ for agent $a_i$ is defined as:
\begin{equation}
\label{eq:features}
\mathcal{X}_i = [x_i^1, x_i^2,\dots, x_i^{k_i}]
\end{equation}
where each element \( x_i^j \) represents a local feature. The number of features $k_i$ may vary across agents, and the internal structure of the individual feature $x_i^j$ can differ as well. 

The agent processes its dataset \( \mathcal{X}_i \) using the local model \( f_i \), which is specifically designed to handle the structure and characteristics of the local data. For each sample, the output of the process is an embedding, defined as:
\begin{equation}
     e \in \mathcal{R}^d
 \end{equation}
where $e$ represents the embedding vector, $\mathcal{R}$ is the set of real numbers, and $d$ denotes the vector space dimension.
Each element of \( \mathcal{X}_i \) is indexed by a position \( t \), which represents either a temporal step or a logical position within the sequence. Therefore, at each step \( t \) the local model produces a local embedding \( e_i^{(t)} \) expressed as:
\begin{equation}
\label{eq:emb}
e_i^{(t)} = f_i(\mathcal{X}_i^{(t)}, w_i)
\end{equation}
where \( w_i \) represents the parameters of the local model specific to each agent and \( \mathcal{X}_i^{(t)} \) is the local features at position \( t \) for agent \( a_i \). Despite the potential diversity in input data and models used, the last layer of each model must produce an embedding with the same structure and dimensions for all agents. Therefore, each local model \( f_i \) can be formally described as a composition of intermediate functions:
\begin{equation}
    f_i = h \circ g_i
\end{equation}
where \( g_i:\mathcal{X} \mapsto z_i\) represents an agent-specific local processing function,  \( z_i \) is an intermediate representation, and \( h : z_i \mapsto e_i \) is a projection function common to all agents. 

\OLshort drives agents to produce local embeddings representing the contextual knowledge extracted from local data. Conceptually, agents sharing the same context \( C \) are expected to produce an identical contextual representation. As a consequence, the agent's objective is to learn to produce, at each step \( t \), local embeddings \( e_i^{(t)} \) that closely resemble those of other agents, using just its local data. Similar embeddings are produced when inputs, though diverse\footnote{It is not similar inputs that lead agents to produce similar embeddings, but rather inputs that, despite their differences, belong to a shared, larger context.}, belong to a shared context that reveals latent correlations. This search for equilibrium among agents is the key to \OLshort that brings out the latent data information of the context. Since each agent acts independently and cannot directly access other agents' embeddings, we can imagine an ideal context embedding \( e_{ctx}^{(t)} \) that represents an optimal convergence point reflecting local embeddings' shared characteristics and providing a unified latent representation of distributed information across the context \( C \). For clarity, \( e_{ctx} \) is a dynamic point in the embedding space that adapts to reflect the patterns and distributions of the context data it represents. A more detailed discussion of how \( e_{ctx} \) is derived is given in Section \ref{sec:osmosis}.

The goal of the training process in \OLshort is, therefore, to ensure that each agent in every iteration produces a local \( e_i^{(t)} \) embedding that aligns with \( e_{ctx}^{(t)} \), while preserving the informational properties of its local data:
    \paragraph{Embedding Alignment Loss Function} 
    the primary objective of the training process is to minimize the distance between each local embedding \( e_i^{(t)} \) and the \( e_{ctx}^{(t)} \) embedding. Given that  each agent only observes its  \( e_i^{(t)} \), we define the alignment loss function as:
    \begin{equation}
        \label{eq:loss-alignment}
        L_{align,i}^{(t)}=d(e_i^{(t)}, e_{ctx}^{(t)})
    \end{equation}
    This function pushes each local embedding  \( e_i^{(t)} \) to converge toward a common point represented by \( e_{ctx}^{(t)} \), reducing the distance in the embedding space and promoting consistency across agents. However, relying solely on alignment with \( e_{ctx}^{(t)} \) may be inefficient. In particular, embedding collapse may occur, where all local embeddings converge to a degenerate point that minimizes the distance to  \( e_{ctx}^{(t)} \) but completely loses the ability to represent meaningful information.
    \paragraph{Local Information Preservation Loss Function} we introduce a secondary loss function to address the risk of embedding collapse, ensuring that local embedding \( e_i^{(t)} \)  preserves the informative properties of local \( \mathcal{X}_i^{(t)} \) data. This prevents the optimization toward \( e_{ctx}^{(t)} \) from resulting in a loss of specificity or degenerate representation.
    A common example of this function is mutual information, which measures the amount of information shared between local embedding and input data:
    \begin{equation}
    \label{eq:loss-preservation}
        L_{pres,i}^{(t)}=-I(e_i^{(t)},\mathcal{X}_i^{(t)})
    \end{equation}

The overall optimization is then defined as a combination of the two loss functions (\ref{eq:loss-alignment}) and (\ref{eq:loss-preservation}):
\begin{equation}
\label{eq:loss}
    L_{total,i}^{(t)}=\lambda L_{align}^{(t)} + (1-\lambda)L_{pres}^{(t)}
\end{equation}
where \( \lambda \) is a hyperparameter that balances the trade-off between context alignment and local information preservation. This formulation allows the definition of different strategies within the \OLshort paradigm by adopting different approaches in designing the loss functions to address these dual objectives effectively and in deriving the optimal \( e_{ctx}^{(t)} \).

The optimization of the total loss function is performed iteratively by updating the local model parameters \( w_i^{(t)} \) based on the gradient of the total loss function \( \nabla_{w_i^{(t)}} \):
\begin{equation}
\label{eq:training}
    w_i^{(t+1)} \leftarrow w_i^{(t)} - \eta \nabla_{w_i^{(t)}} L_{total, i}^{(t)}
\end{equation}

where \( \eta \) denotes the learning rate. The training process is organized into multiple epochs, each performing a complete pass through an agent's data \( \mathcal{X}_i \), where local model parameters are iteratively updated using data batches. 

\subsection{Intrinsic Data Order and Contextual Alignment}
For \OLshort to be meaningfully applied, the local data must possess an inherent logical order within the context \( C \), which must be maintained consistently during both training and inference to ensure a coherent alignment. This is particularly evident in time series data, where the natural arrangement follows a defined temporal sequence. At each time step, the data from all agents collectively represent the overall shared information of the context \( C \) to which they belong, effectively capturing a snapshot of its temporal dynamics. Similarly, an order can also emerge in other data types as a logical or structural relationship, attributing clear meaning to their arrangement.
Considering the local feature vector \( \mathcal{X}_i \) of an agent \( a_i \), each feature \( x_i^{(j)} \) is structured as an ordered sequence of elements:
\begin{equation}
    x_i^{(j)} = (x_i^{(j,1)}, x_i^{(j,2)}, \dots, x_i^{(j,T)})
\end{equation}
Each temporal or logical index \( t \) of the sequence, given the set of agents and \( k_i \) as the number of features for agent \( a_i \), corresponds to a context tuple:
\begin{equation}
    \mathcal{T}^{(t)} = (\{x_1^{(0,t)}\}_{j=1}^{k_1}, \{x_2^{(1,t)}\}_{j=1}^{k_2}, \dots, \{x_n^{(n,t)}\}_{j=1}^{k_n})
\end{equation}
This tuple represents the set of local data for agents at the exact logical location \( t \) within their respective sequences. In the case of using Recurrent Neural Networks (RNN), where the concept of sliding windows is employed as model input, the feature representation \( x_i^{(j)} \) represents a specific window corresponding to a given feature.

\subsection{Osmosis}
\label{sec:osmosis}
The concept of osmosis lies at the core of \OLshort, enabling the determination of the optimal \( e_{ctx}^t \) context embedding as a synthesis of local embeddings produced by agents. Using an \textit{osmotic strategy} $\Omega$, local embeddings are analyzed to detect latent patterns and structures within the distributed data, providing a consistent and meaningful global representation. The osmotic strategy is versatile and can be tailored to specific application needs or data properties.
Thus, we formulate the problem as determining the point in the representation space that minimizes the overall distance between local embeddings:
\begin{equation}
\label{eq:osmotic-strategy}
\Omega : e_{ctx}^{(t)} = \arg \min_{e \in E} \sum_{i \in A} d(h(g_i(X_i^{(t)})), e)
\end{equation}
where \( d \) is a distance function (e.g., Euclidean distance or cosine similarity) and \( E \) represents the space of possible context representations.

\subsection{Shared Latent Information}
\label{sec:shared-latent-information}
Local embeddings converging to a context embedding hold a latent, shared representation of information of the context that naturally emerges as a result of the distributed \OLshort intrinsic process.
During training, agents learn local representations from their data, but these representations are not isolated; they are influenced by context constraints, driven by the push to achieve a common balance point among embeddings. 
Although mediated solely by local information, this process of iterative knowledge exchange identifies recurrent patterns that synthesize unspoken correlations among local agent data. 

These patterns reflect hidden structures and shared relationships within the context that would not be observable by analyzing local data in isolation.
This information represents the common latent state that connects distributed data, incorporating temporal dynamics, logical relationships, or shared semantic structures. Correlation among local data is essential for its emergence: highly correlated data enable context embedding to synthesize a coherent and meaningful space beyond the simple aggregation of local representations.
For this reason, \OLshort must be applied in a specific distributed context where the distributed data are strongly interconnected and interdependent. This interaction ensures that local representations are not fragmented or inconsistent but can contribute significantly to constructing a meaningful and stable context embedding.

\subsection{Clustering and sub-contexts}
The correlation among agents' local data is essential for consistent alignment of representations. However, correlation varies since some agents are strongly correlated, others partially, and some weakly.
When correlation is insufficient, the local embedding alignment algorithm fails to converge to a unified context representation for all agents.

Let \( e_i^{(t)} \) be the local embedding produced by agent \( a_i \) at position \( t \) and  \( e_{ctx}^{(t)} \) the aggregate optimal context embedding. A similarity function \( s \) quantifies the degree of similarity between local embeddings. If two agents share a significant correlation, their similarity exceeds a defined threshold \( \tau \):
\begin{equation}
    s(e_i^{(t)},  e_j^{(t)}) \ge \tau
\end{equation}
During training, this behavior enables the dynamic formation of agent clusters, referred to as \textit{sub-context}. Such clusters naturally emerge as groups of agents whose local embeddings show a reduced distance and align along a common trajectory toward a shared context embedding:
\begin{equation}
    e_{ctx}^{(t,k)} = \arg \min_{e \in E} \sum_{i \in A_k} d(e_i^{(t)}, e)
\end{equation}
where \( A_k \) denotes the set of agents within the sub-context \( C_k \subseteq C \), and \( e_{ctx}^{(t,k)} \) denotes the optimal context embedding for the sub-context \( C_k \) at position \( t \).
This optimization improves the alignment of agents within each cluster while reducing the interference from unrelated agents. Agents belonging to multiple clusters contribute to multiple context representations, acting as semantic bridges between different groups.

%% file: tex/4-architecture.tex
\section{Architecture}
\label{sec:architecture}
\begin{figure}[t!]
    \centering
    \includegraphics[width=1\linewidth]{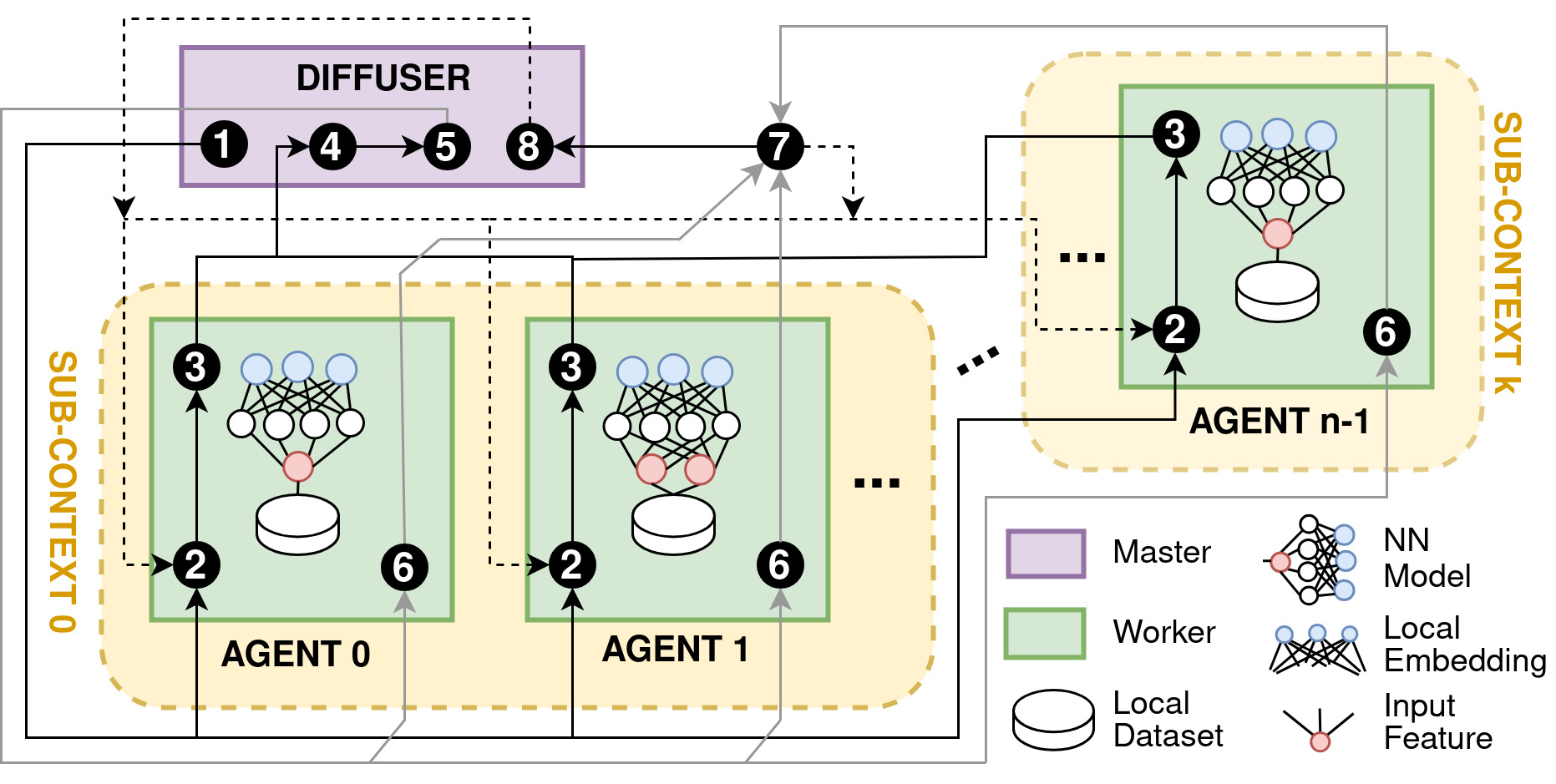}
    
    \caption{\OLshort architecture.} 
    \label{fig:architecture}
\end{figure}

The architecture of \OLshort shown in Fig.~\ref{fig:architecture} adopts a worker-master scheme, where a distributed network of \textit{agents} in worker instances processes local data and a central master server,  \textit{diffuser}, orchestrates the global alignment by aggregating local embeddings into a shared global representation. It indeed balances client autonomy with the construction of a shared global representation through an iterative, dynamic \textit{diffusion} of information via osmosis.

\paragraph{Agents}
are independent workers who process local data using models designed according to the nature of available features. Each agent produces dense,  compact local representations, i.e., local embeddings, while preserving the temporal or logical order of the data throughout processing. Despite their autonomy, agents must align with a shared global representation to ensure consistency across local representations.

\paragraph{Diffuser}
serves as a central orchestrator, aggregating local embeddings from agents into an optimal global representation using \textit{osmotic strategy}. This strategy captures latent correlations among distributed data, reflecting common patterns and shared structures. In addition, the diffuser dynamically identifies sub-contexts by clustering agents with similar local embeddings, managing agents membership across sub-contexts, and deriving multiple global embeddings without the agents' awareness.

The training process in \OLshort consists of iterative epochs, each consisting of key steps orchestrated by the diffuser. In each epoch, the diffuser and agents collaboratively refine local embeddings, achieving convergence to a global embedding. The process includes the following steps: 
\begin{enumerate}[label=\protect\numbercircle{\arabic*}]
\item \textit{Initialization}: the diffuser begins the epoch by distributing initial reference parameters to the agents;
\item\textit{Forward pass}: each agent processes a batch of ordered local data by passing it through its model, generating a dense local embedding that captures key features and relationships without updating model parameters;
\item\textit{Embedding sharing with the diffuser}: agents transmit their generated local embeddings to the diffuser, which aggregates them for central processing;
\item\textit{Osmotic strategy}: the diffuser  processes all local embeddings to calculate optimal context or sub-context embeddings;
\item\textit{Diffusion of the context embedding}: the diffuser distributes context or sub-context embeddings to agents, providing a reference for local optimization;
\item\textit{Loss calculation and backpropagation}: each agent calculates the total loss (Eq. \ref{eq:loss}) and performs backpropagation; local model parameters are updated using the calculated gradient, progressively improving alignment with the context embedding;
\item\textit{Batch iteration}: repeat from step \numbercircle{2} until all agent batches are processed;
\item\textit{Clustering}: the diffuser groups agents into sub-contexts based on the similarity of their local embeddings, either at the end of each epoch or as dictated by a specific policy.
\end{enumerate}
The training process repeats from step \numbercircle{2} until the final epoch is reached, ensuring iterative refinement.

%% file: tex/5-evaluation-setup.tex
\section{Evaluation Setup}
\label{sec:setup}
This section details the \OLshort experimental setup, datasets, and evaluation metrics.
\subsection{Experimental Setup}
We employed Python 3.10 and \texttt{PyTorch} library to develop a simulator for implementing \OLshort.
We used a small GRU\footnote{\url{https://pytorch.org/docs/stable/generated/torch.nn.GRU.html}} (Gated Recurrent Unit) neural network model for each agent with a hidden dimension of \num{20}. We added a fully connected Linear\footnote{\url{https://pytorch.org/docs/stable/generated/torch.nn.Linear.html}} layer as the final network component to generate the local embedding, fixed at a size of \num{5} for all agents (model details are in Table~\ref{tab:gru_model}). We trained the model using the Adam optimizer\footnote{\url{https://pytorch.org/docs/stable/generated/torch.optim.Adam.html}} with a learning rate $\eta=0.001$. We employed Mean Squared Error (MSE) as the distance function for the embedding alignment loss function~(\ref{eq:loss-alignment}), while mutual information, approximated with a contrastive loss function scaled by a temperature parameter of \num{0.1}, ensures local information preservation~(\ref{eq:loss-preservation}). We set the sliding window to build sequences of length \num{10} and organized the input data into batches of size \num{50}, with samples ordered chronologically by timestamp. We configured the diffuser with an \textit{osmotic strategy} that minimizes the distances between local embeddings defined in Eq.~\ref{eq:osmotic-strategy}. We also enabled clustering to dynamically identify sub-contexts by comparing a set of \num{20} random local embeddings per agent every two epochs using a similarity threshold $\tau=0.97$.
\begin{table}[t!]
\centering
\caption{GRU Model Architecture.}
\begin{tabular}{lcc}
\toprule
\textbf{Layer} & \textbf{Parameters} & \textbf{Output Shape} \\
\midrule
Input & - & $[\text{batch\_size}, \text{seq\_len}, \text{n\_features}]$ \\

GRU & $60\times\text{n\_features}+1320$ & $[1, \text{batch\_size}, 20]$ \\

Linear &  $105$ & $[\text{batch\_size}, 5]$ \\

\bottomrule
\end{tabular}
\label{tab:gru_model}
\end{table}



\subsection{Evaluation Metrics}
We assessed \OLshort using three key metrics to measure representation alignment and generalization.
     \paragraph{Embeddings similarity} estimates the similarity between local embeddings using a modified \textit{cosine similarity} function, designed to accentuate stronger correlations and dampen weaker ones. Formally, the similarity between two embeddings \( e_a \) and \( e_b \), is defined as:
    \begin{equation}
    \label{eq:embedding-similarity} 
    s(e_a, e_b) = \text{sign}(s_{c}(e_a, e_b)) \cdot |s_{c}(e_a, e_b)|^\beta 
\end{equation} 
    where $s_{c}$ represents the standard cosine similarity\footnote{\url{https://pytorch.org/docs/stable/generated/torch.nn.CosineSimilarity.html}}, and $\beta>1$ is a hyperparameter that amplifies significant alignments.
    \paragraph{Context accuracy} measures the average similarity between local embeddings generated by agents within the same context. Using a cosine similarity function $s$, normalized between \num{0} and \num{1}, context accuracy is calculated as:
        \begin{equation}
        \label{eq:accuracy}
        accuracy_{ctx} = \frac{2}{n(n-1)T} \sum_{t=0}^{T-1} \sum_{i=0}^{n-2} \sum_{j=i+1}^{n-1} s(e_i^{(t)}, e_j^{(t)})
        \end{equation}
\paragraph{Context loss} evaluates the average total loss of individual agents across both training and test datasets:
        \begin{equation}
        \label{eq:loss-metric}
        loss_{ctx} = \frac{1}{nT} \sum_{t=0}^{T-1} \sum_{i=0}^{n-1} L_{total, i}^{(t)}
        \end{equation}
Note that when sub-contexts are formed, \textit{context accuracy} and \textit{context loss} are calculated as the average of these respective metrics across individual sub-contexts.
 
\subsection{Datasets}
We designed synthetic time series data with a structured and interpretable format to highlight the unique capabilities of \OLshort, ensuring a clear evaluation of its contextual representation and alignment processes, which are not explicitly captured in existing datasets. This design ensures precise isolation of the \OLshort process, reducing ambiguities and facilitating a deeper understanding of the expected outcomes.
Two distinct contexts were constructed, each providing agents with training (\num{1000} samples per feature) and testing (\num{200} samples per feature) datasets, including one or more features with iterative behaviors.

\begin{figure}[!t]
    \centering
    \includegraphics[width=\linewidth]{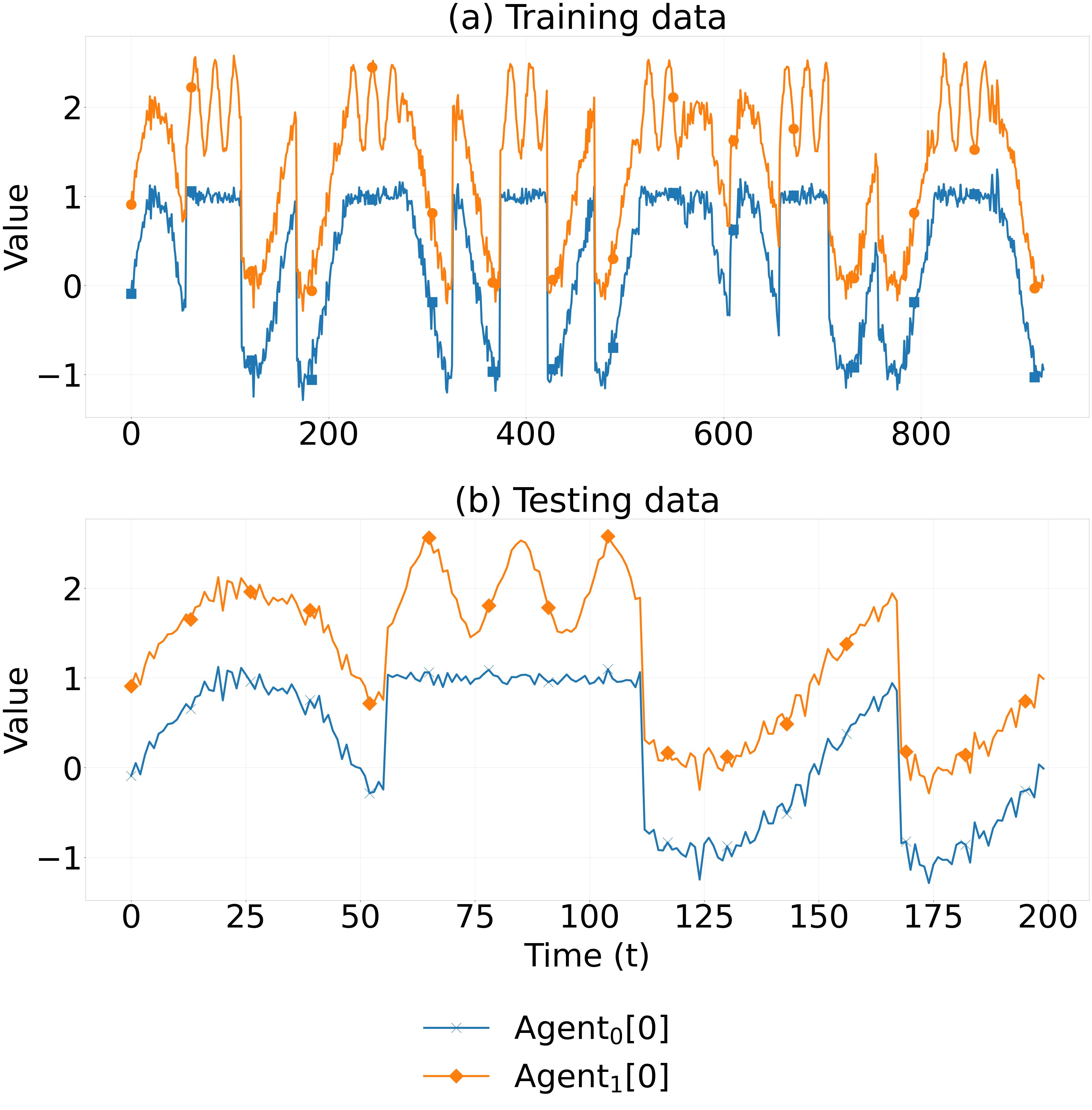}
    \caption{Simple context. Training (a) and testing (b) data of two agents with correlated features.}
    \label{fig:simple-context-data}
\end{figure}

\paragraph{Simple context}\label{p:simple-context} comprises two agents, $Agent_0$ and $Agent_1$, each with a dataset containing a single feature. The data show a strong correlation (Fig. \ref{fig:simple-context-data}), indicating positive synchronization in their time series values. $Agent_0$ typically mirrors the behavior of $Agent_1$, but during specific phases, while $Agent_1$ remains stationary, $Agent_0$ demonstrates oscillatory behavior. From $Agent_0$'s perspective, these appear unstable. However, at the global context level, they can be considered stationary.
\paragraph{Complex context}\label{p:complex-context} builds upon the simple context using five agents (Fig. \ref{fig:experiment-2-data}), each showing distinct characteristics:
\begin{itemize}
    \item $Agent_0$ and $Agent_1$ maintain the same features as in the simple context with an added offset in their data, enabling testing \OLshort's invariance to constant data shifts;
    \item $Agent_2$ and $Agent_3$ each have a single feature, with their data exhibiting a strong negative correlation. Specifically, when one agent's values increase, the other's decrease, reflecting an inverse relationship.; 
    \item $Agent_4$ has two features, each partially correlated with those of $Agent_2$ and $Agent_3$. The first feature shares a moderate positive correlation discretized between values \num{0} and \num{10}, while the second feature shows a partial positive correlation with $Agent_2$ and a partial negative correlation with $Agent_3$. Additionally, the second feature introduces a sinusoidal behavior, further diversifying its dynamics.
\end{itemize}

%% file: tex/6-experimental-results.tex
\section{Experimental Results}
\label{sec:experiment}
This section examines the \OLshort effectiveness through experiments, focusing on the quality of latent information shared via embeddings, training convergence, and the dynamic formation of sub-contexts.
\subsection{Latent Representation and Training Convergence}
\begin{figure}[t!]
    \centering
    \includegraphics[width=0.46\linewidth]{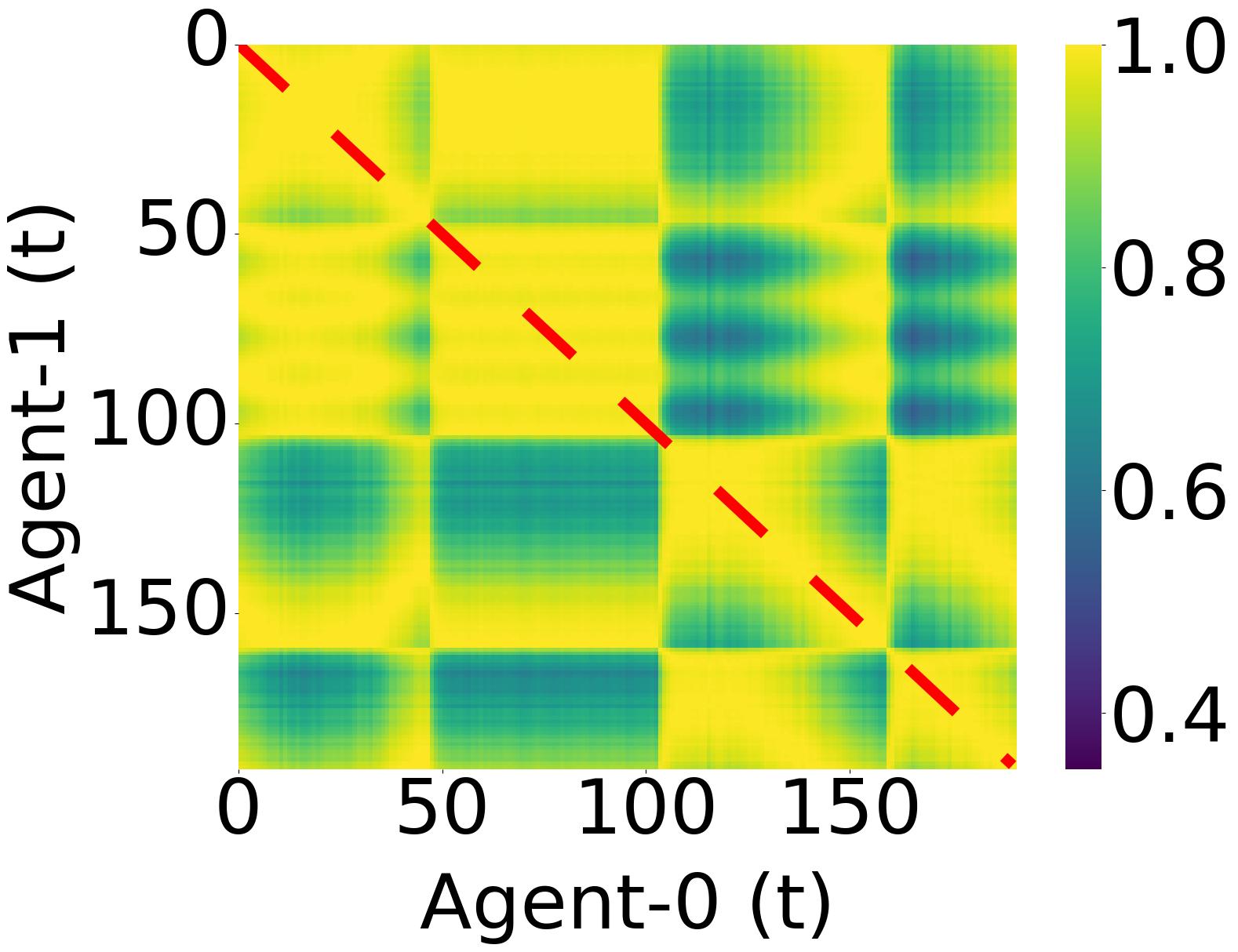}
    \caption{Embedding similarity matrix in the simple context.} 
    \label{fig:experiment-0-similarity}
\end{figure}

In the first experiment, we trained $Agent_0$ and $Agent_1$ in the simple context (\ref{p:simple-context}) for five epochs, evaluating their models on the test dataset and achieving an accuracy of \num{0.9984} at the last epoch.
The local embeddings are evaluated using a similarity matrix, as shown in Fig.~\ref{fig:experiment-0-similarity} constructed based on Eq.~\ref{eq:embedding-similarity} with $\beta=2$, where each embedding produced by one agent at a given step $t$ is compared with all embeddings generated by the other agents.

The similarity matrix (Fig.~\ref{fig:experiment-0-similarity}) provides key insights into the evolution of embeddings and the learning process throughout the evaluation period: 1) the \textit{main diagonal} serves as a primary indicator of convergence, reflecting the similarity between embeddings generated by agents at the same logical or temporal position. 2) the overall \textit{matrix structure} unveils hidden patterns, illustrating how \OLshort drives agents toward representations that not only capture temporal dependencies within local data but also integrate them into a coherent global context.
\begin{figure}[!t]
    \centering
    \includegraphics[width=\linewidth]{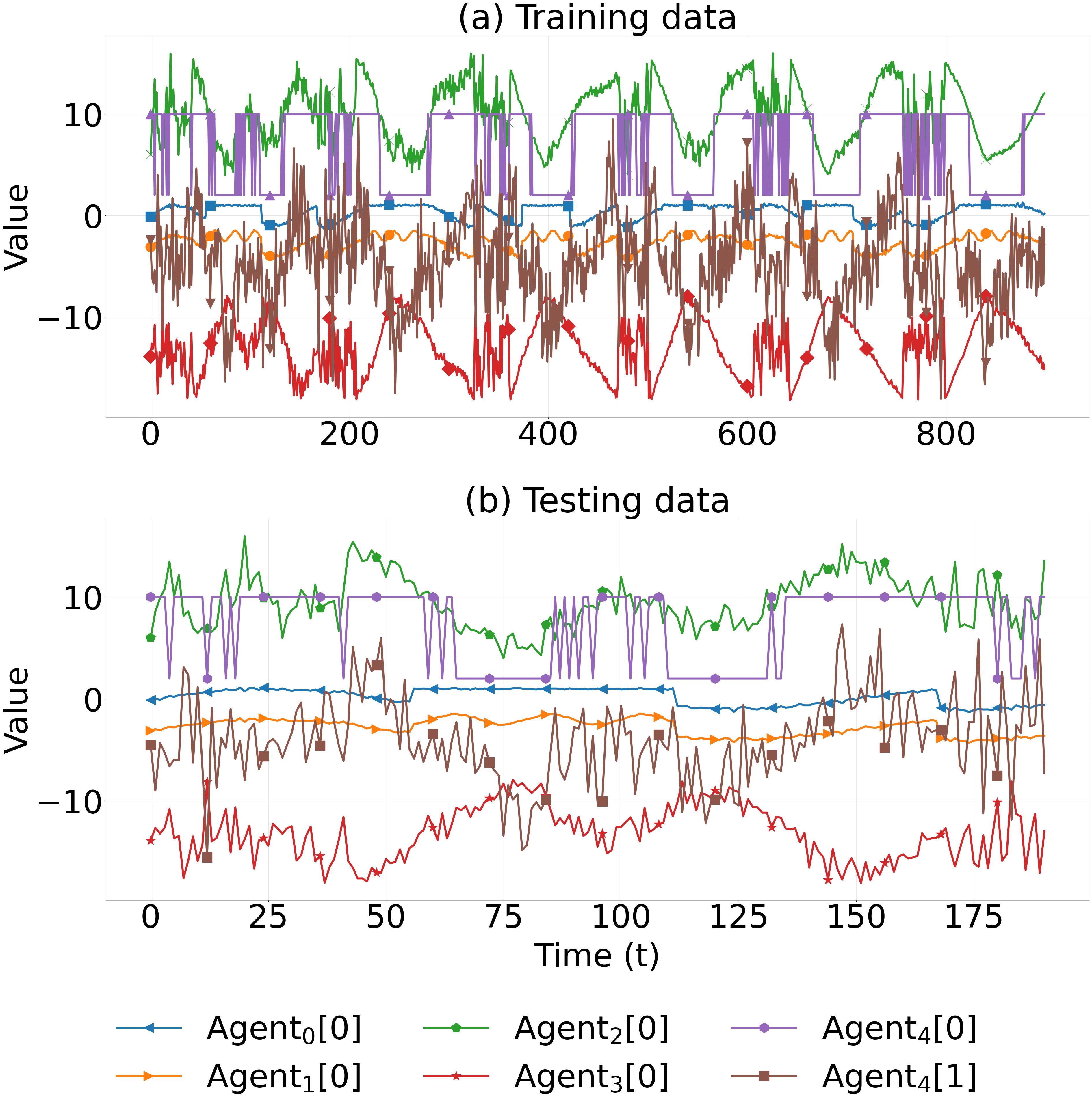}
    \caption{Complex context. Training (a) and testing (b) data of five agents. The data are structured to highlight the correlation between the two groups: one formed by $Agent_0$ and $Agent_1$, and the other by $Agent_2$--$Agent_4$.}
    \label{fig:experiment-2-data}
\end{figure}

The similarity along the diagonal demonstrates how \OLshort generates consistent representations at the same temporal positions despite independent training processes for the two agents. This result also confirms the effectiveness of the iterative alignment process, as local embeddings converge toward a shared representation: from only local data, both $Agent_0$ and $Agent_1$ independently derive the same contextual information.
Moreover, the matrix region within the interval $[50,100]$ for both agents highlights \OLshort's ability to recognize this phase as stationary, regardless of the oscillatory behavior of $Agent_0$'s feature.

\begin{figure*}[!t]
    \centering
    \begin{subfigure}[b]{0.157\linewidth}
        \includegraphics[width=\linewidth]{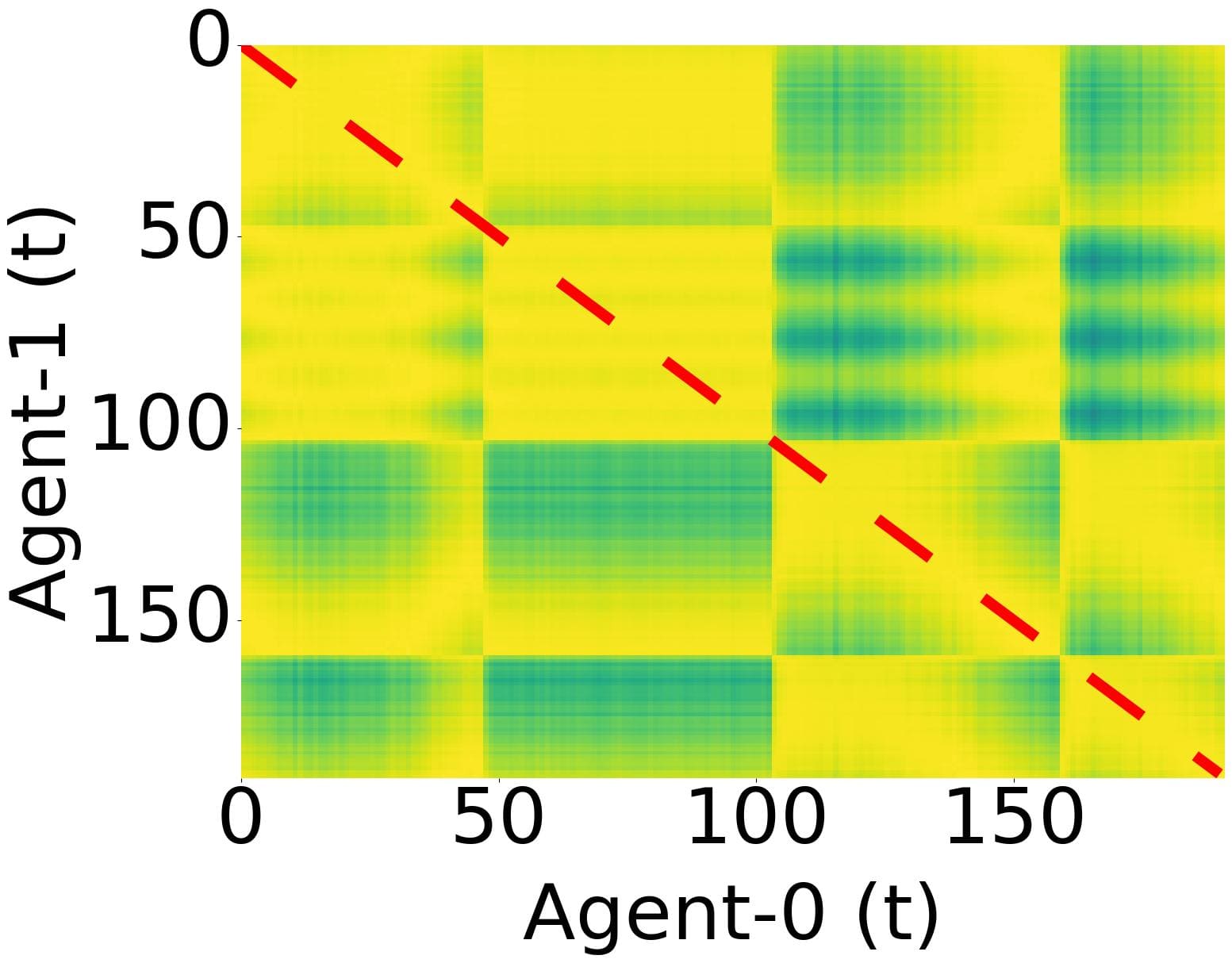}
    \end{subfigure}
    \begin{subfigure}[b]{0.157\linewidth}
        \includegraphics[width=\linewidth]{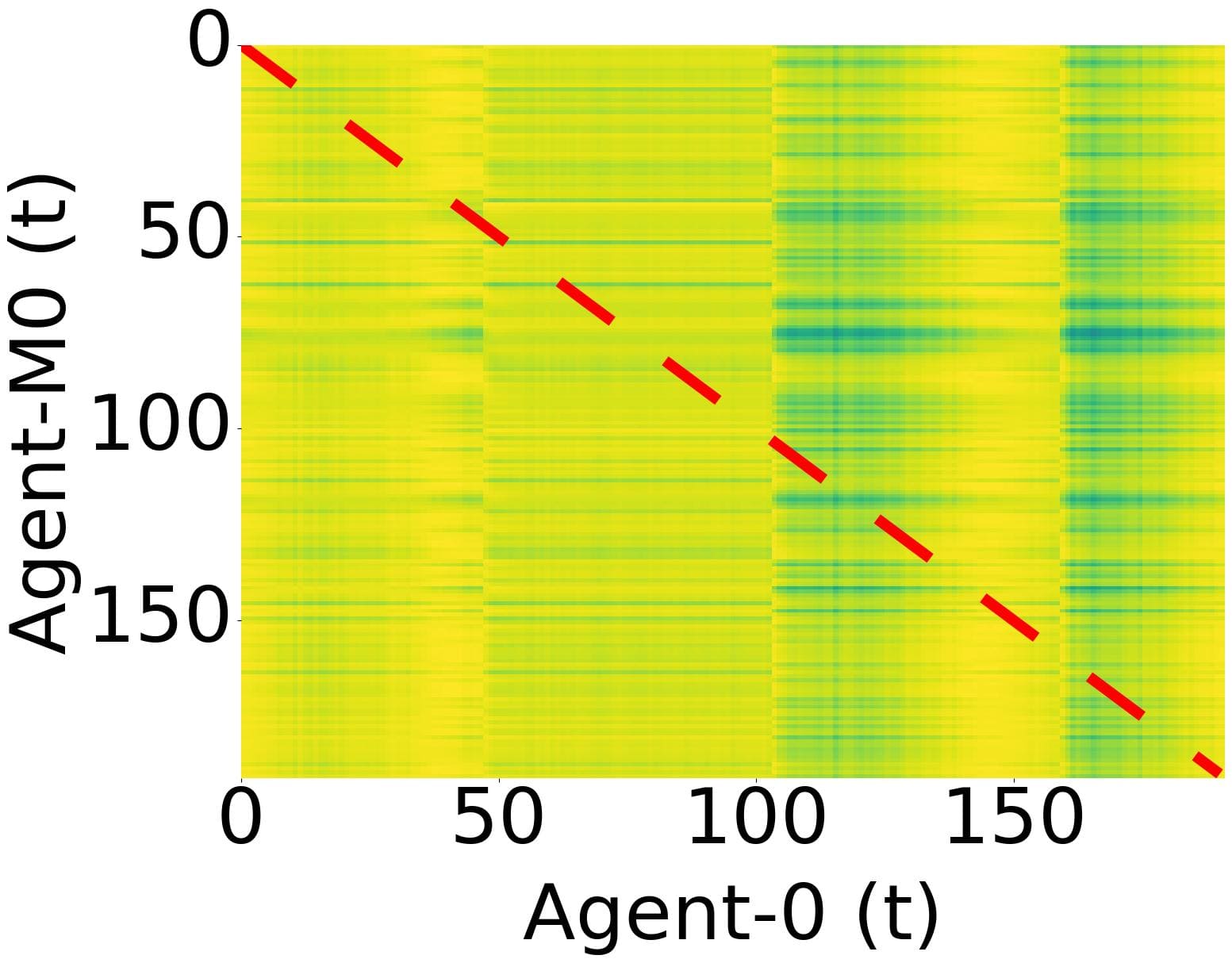}
    \end{subfigure}
    \begin{subfigure}[b]{0.157\linewidth}
        \centering
        \includegraphics[width=\linewidth]{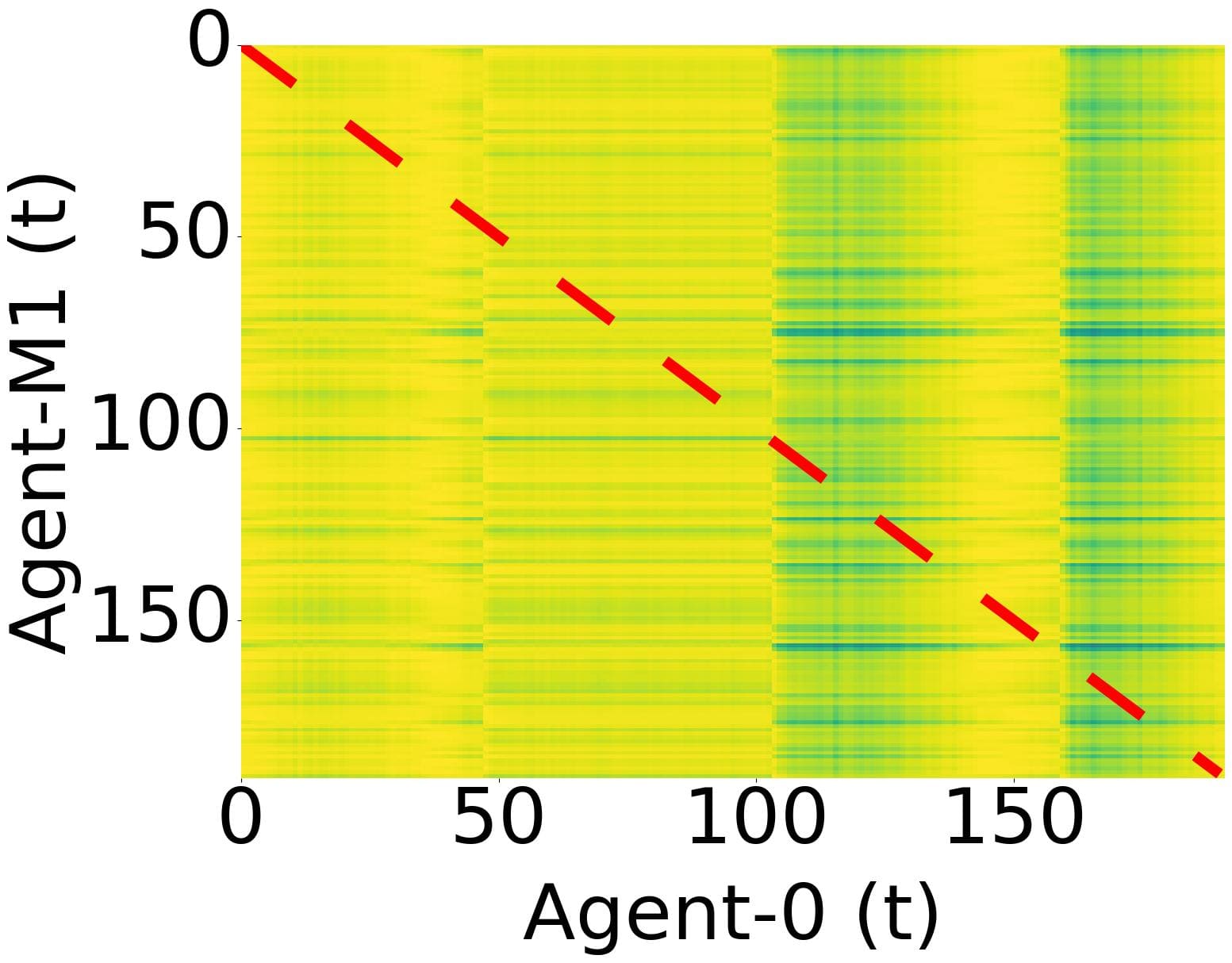}
    \end{subfigure}
        \begin{subfigure}[b]{0.157\linewidth}
        \includegraphics[width=\linewidth]{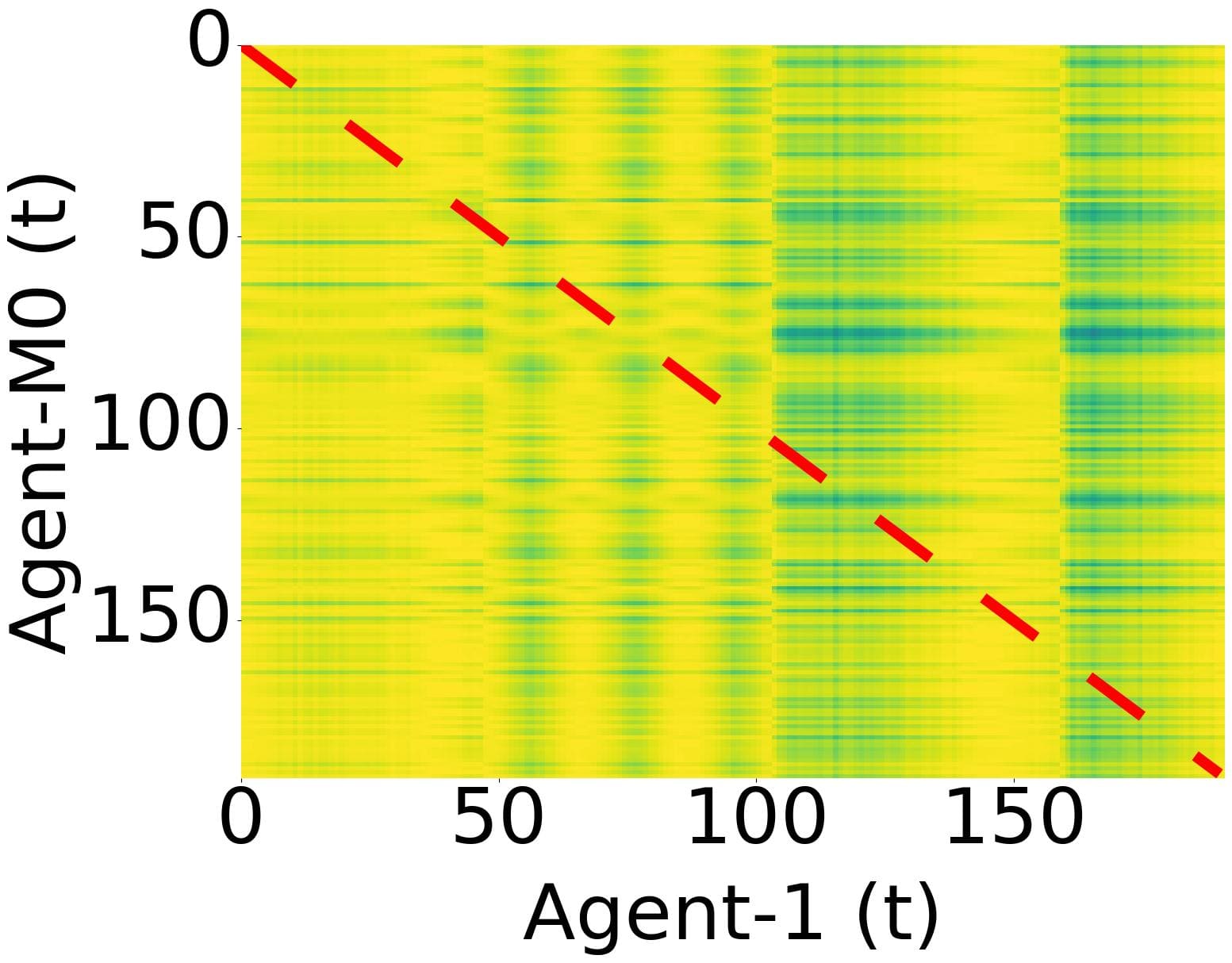}
    \end{subfigure}
    \begin{subfigure}[b]{0.157\linewidth}
        \includegraphics[width=\linewidth]{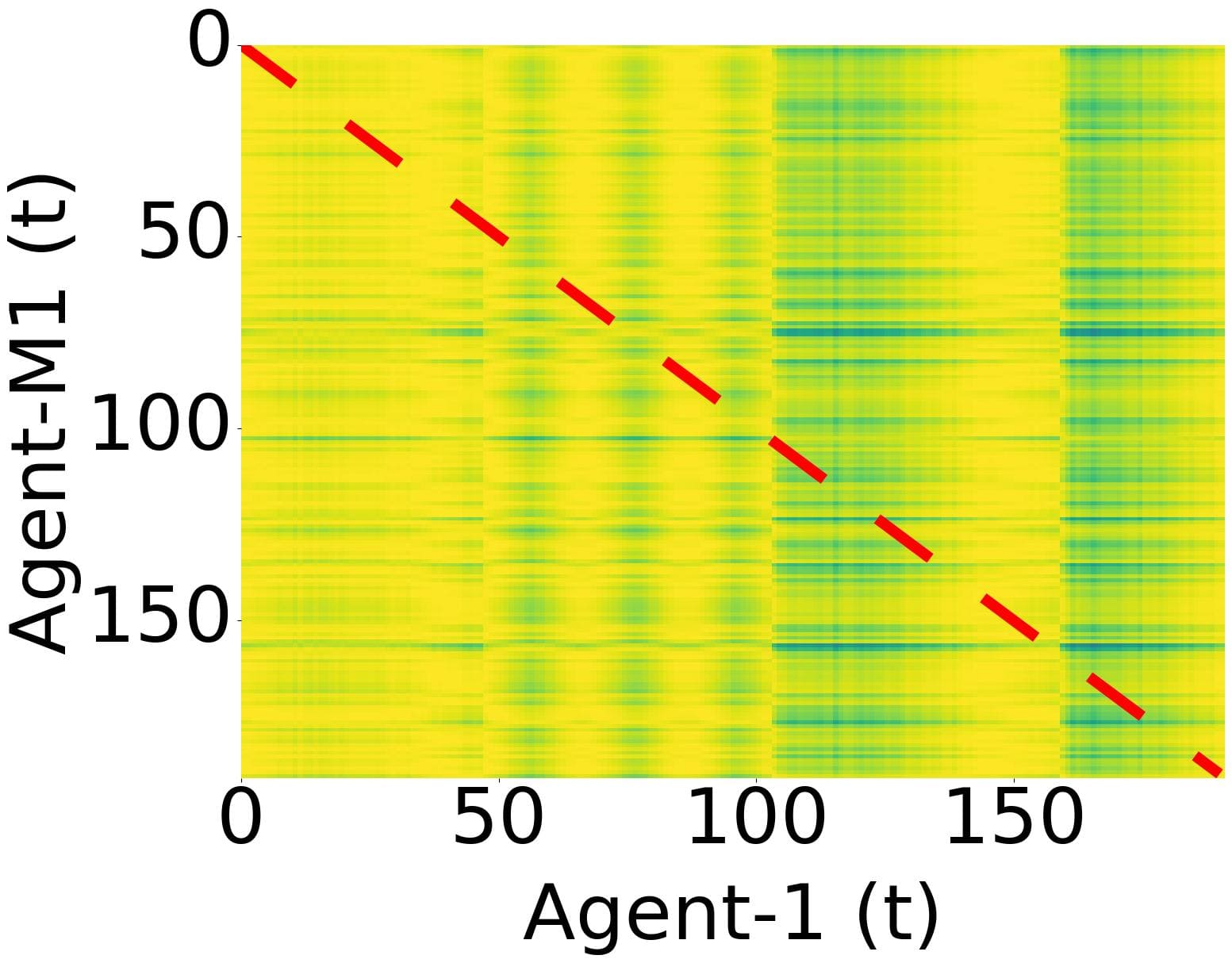}
    \end{subfigure}
    \begin{subfigure}[b]{0.172\linewidth}
        \centering
        \includegraphics[width=\linewidth]{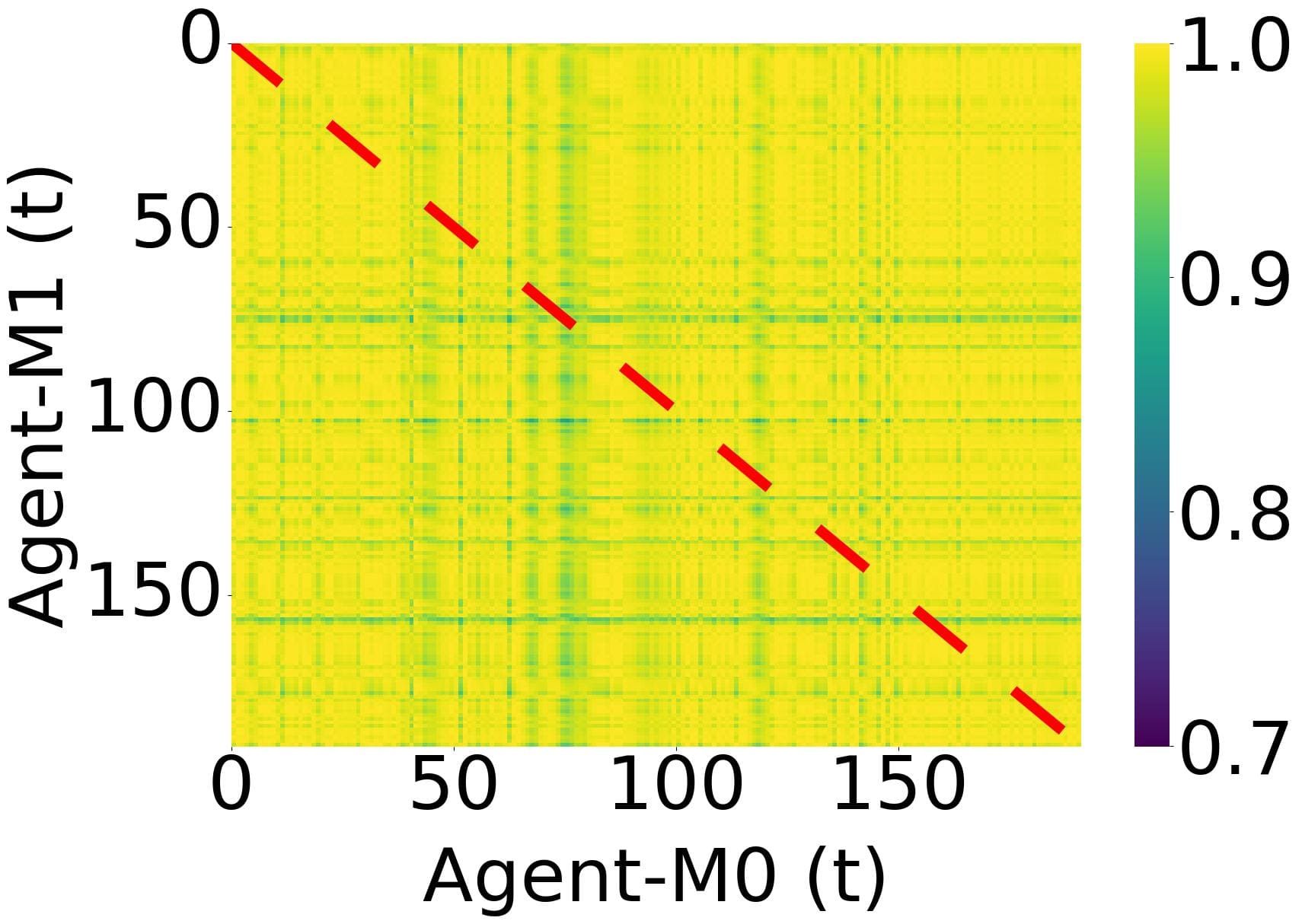}
    \end{subfigure}
    \caption{Embedding similarity matrices in the simple context with two additional misleading agents.}
    \label{fig:convergence-with-noise}
\end{figure*}
\begin{figure*}[t!]
    \centering
    \begin{subfigure}[b]{0.19\linewidth}
        \includegraphics[width=\linewidth]{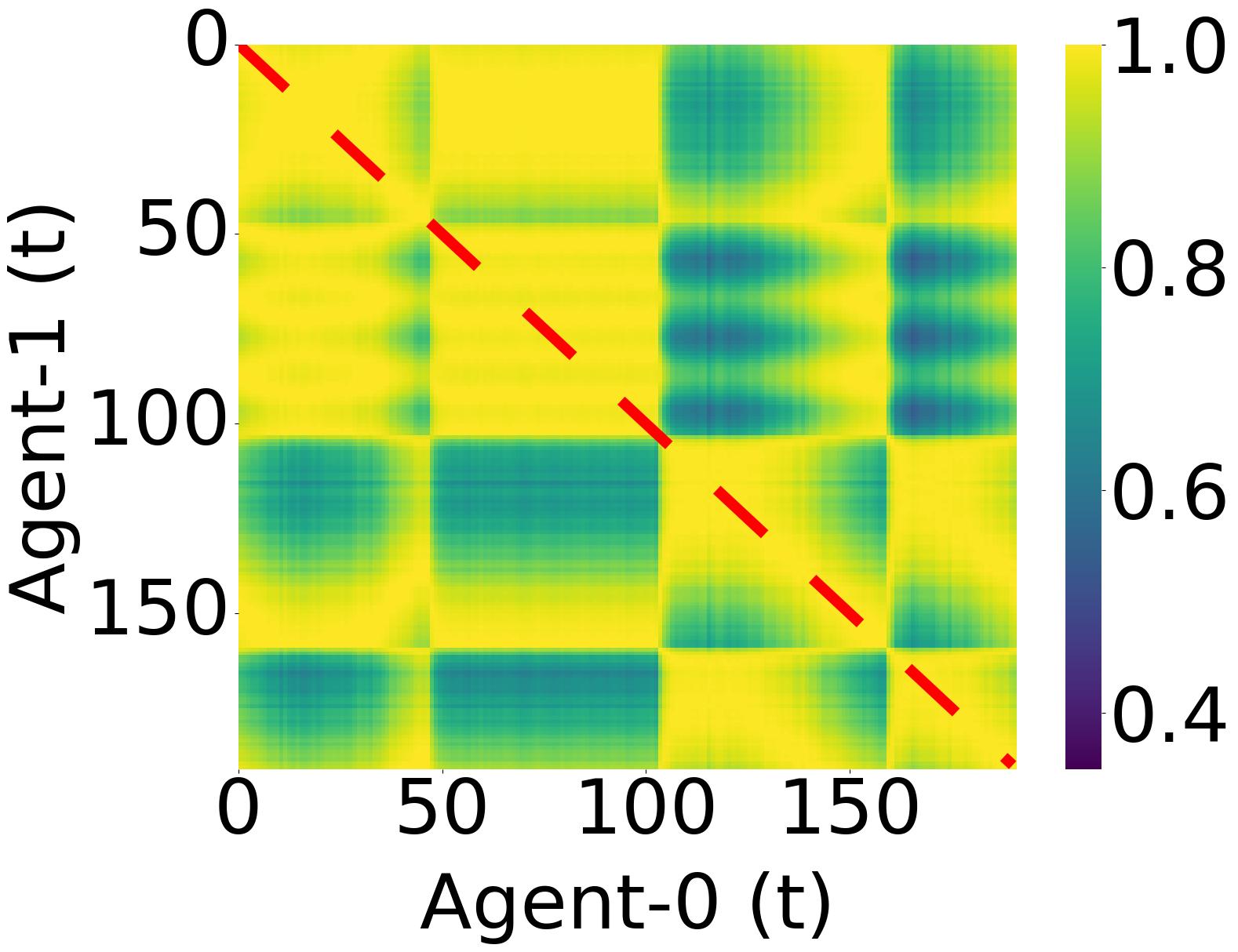}
    \end{subfigure}
    \begin{subfigure}[b]{0.19\linewidth}
        \includegraphics[width=\linewidth]{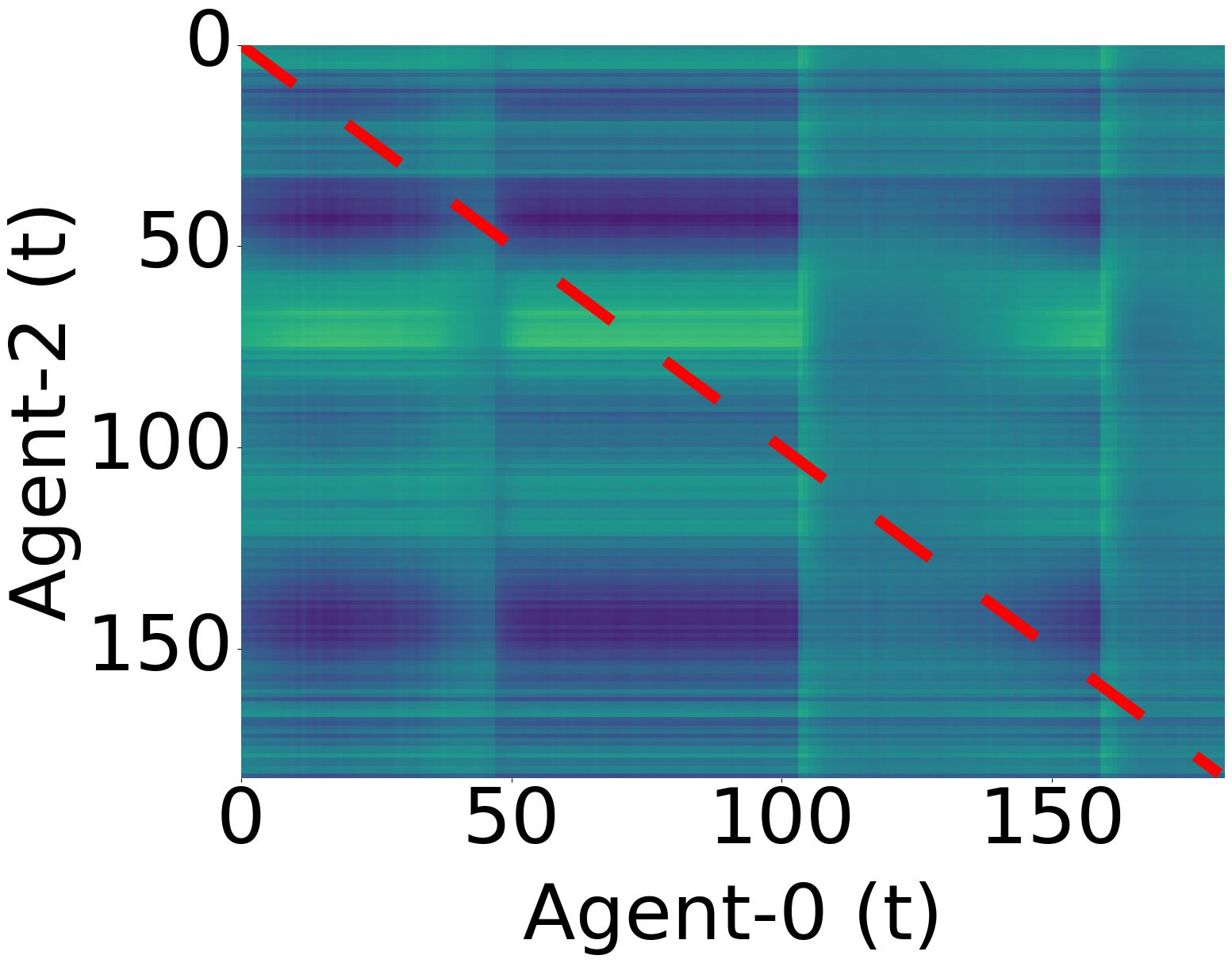}
    \end{subfigure}
    \begin{subfigure}[b]{0.19\linewidth}
        \centering
        \includegraphics[width=\linewidth]{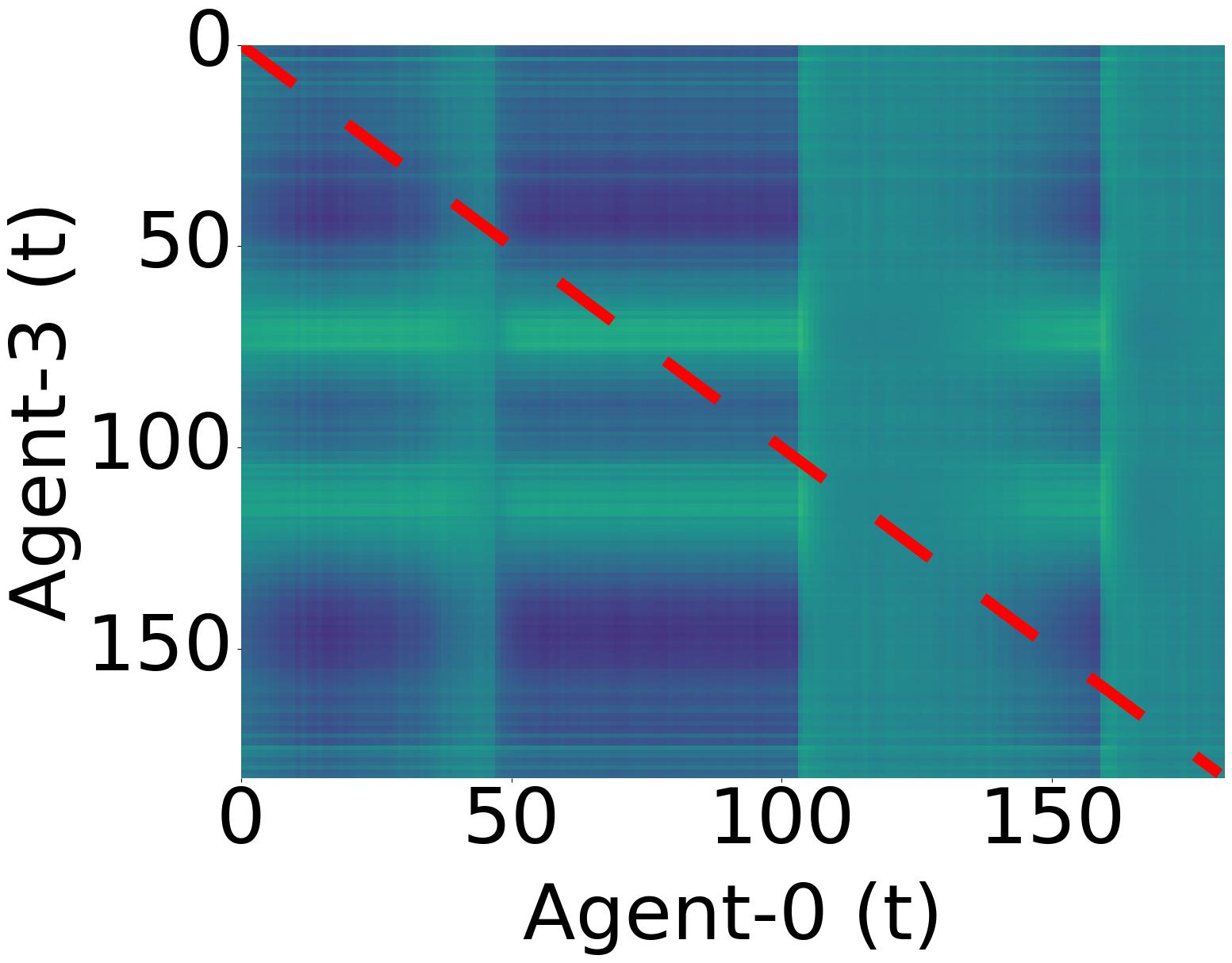}
    \end{subfigure}
        \begin{subfigure}[b]{0.19\linewidth}
        \includegraphics[width=\linewidth]{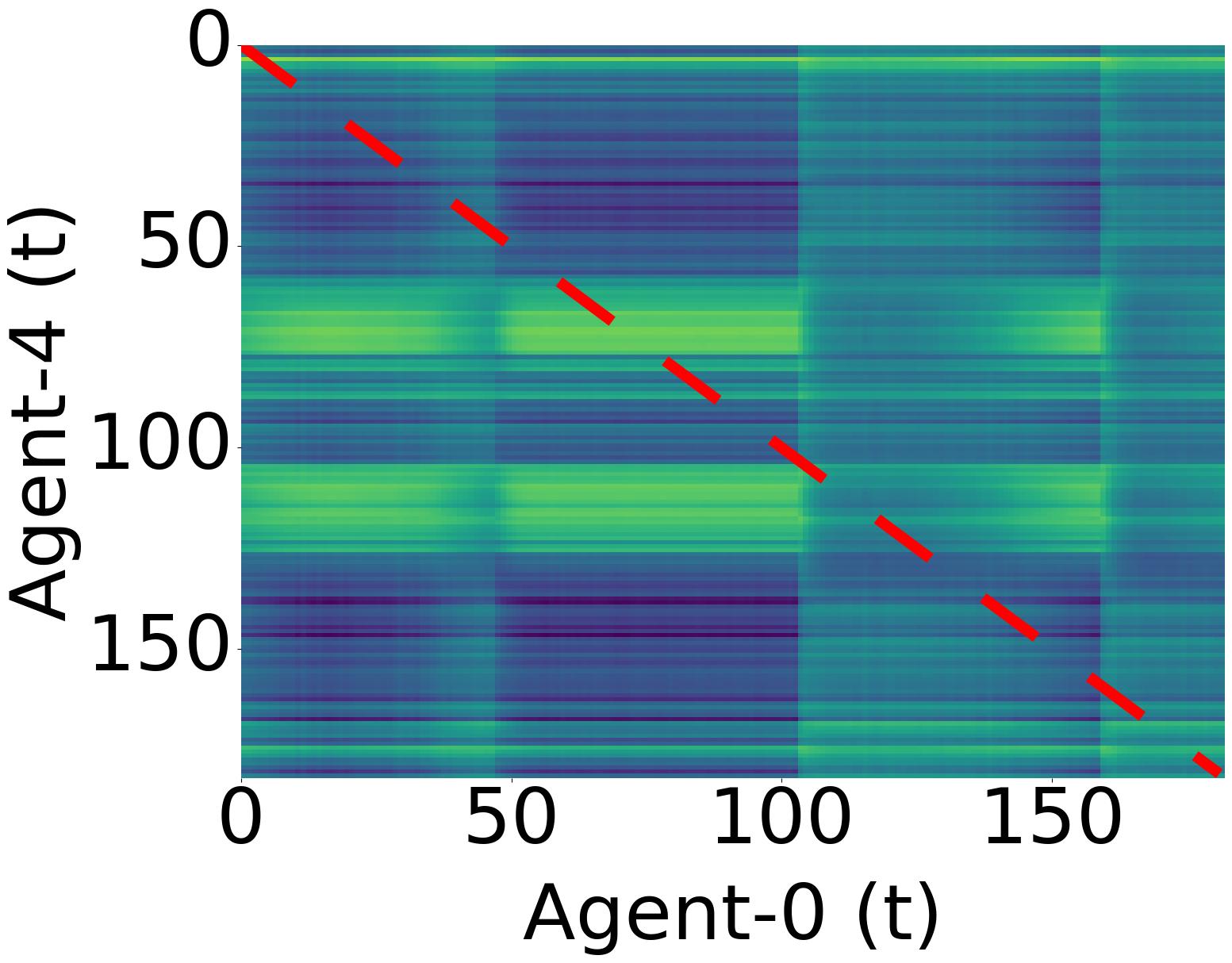}
    \end{subfigure}
    \begin{subfigure}[b]{0.19\linewidth}
        \includegraphics[width=\linewidth]{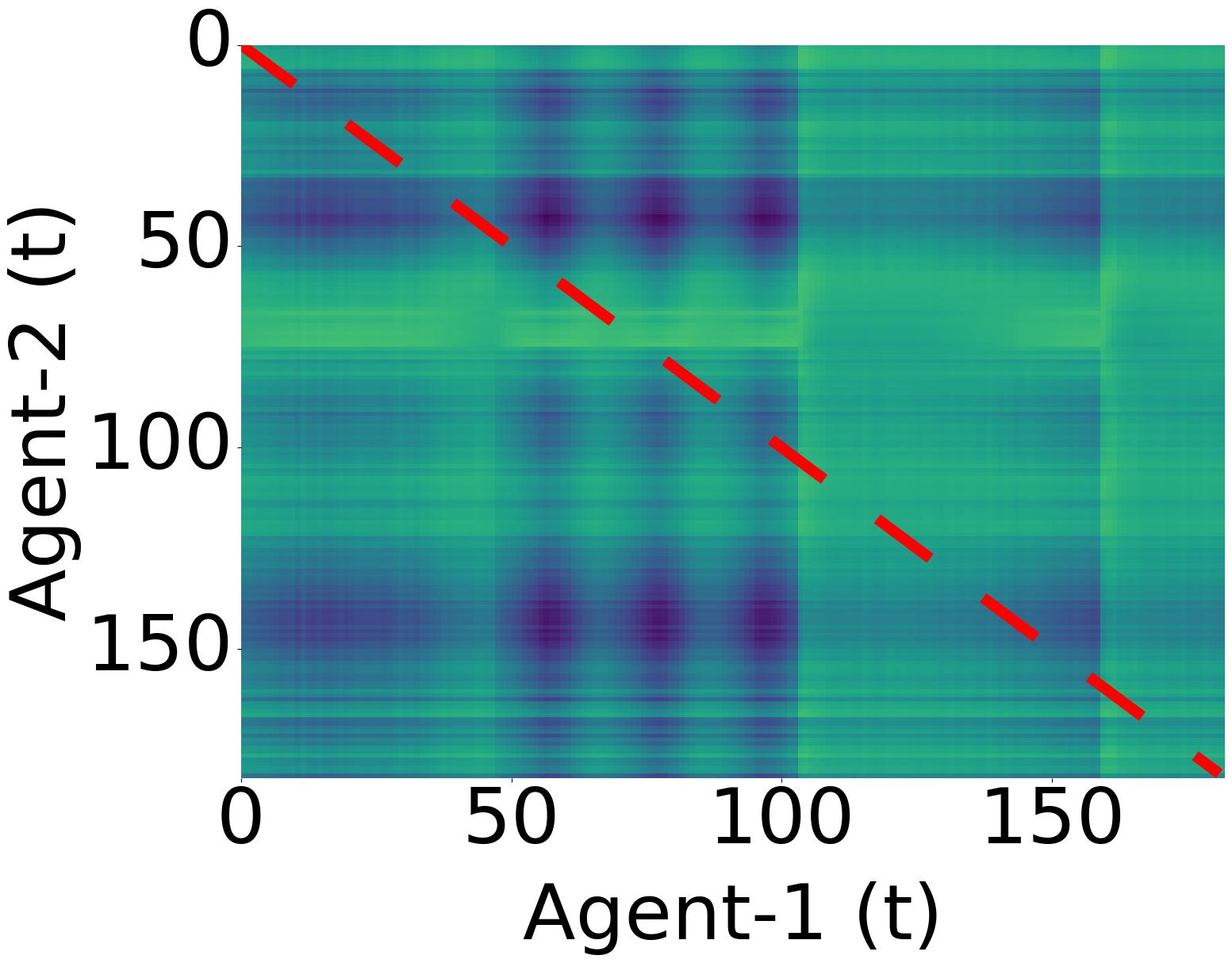}
    \end{subfigure}
        \begin{subfigure}[b]{0.19\linewidth}
        \includegraphics[width=\linewidth]{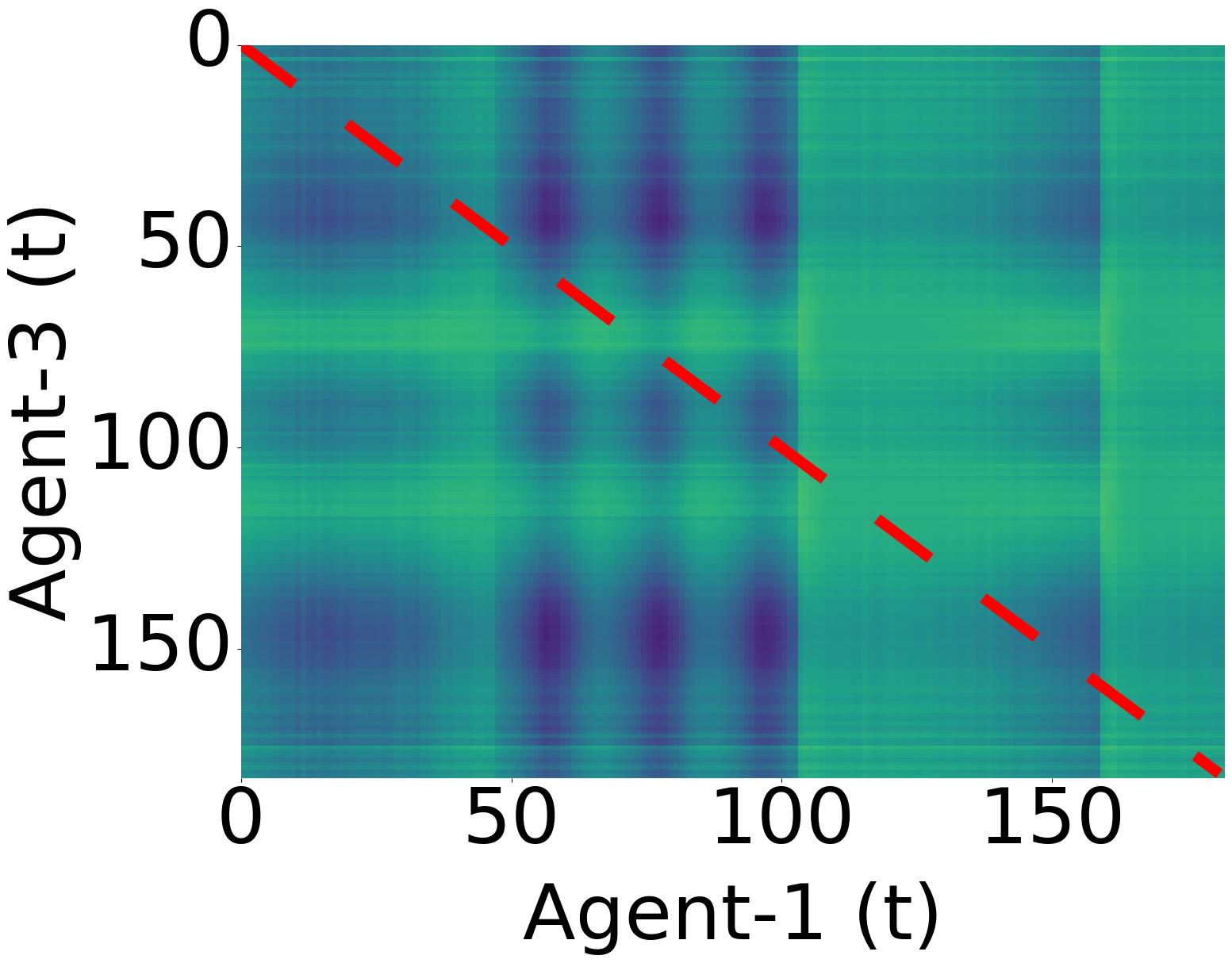}
    \end{subfigure}
    \begin{subfigure}[b]{0.19\linewidth}
        \includegraphics[width=\linewidth]{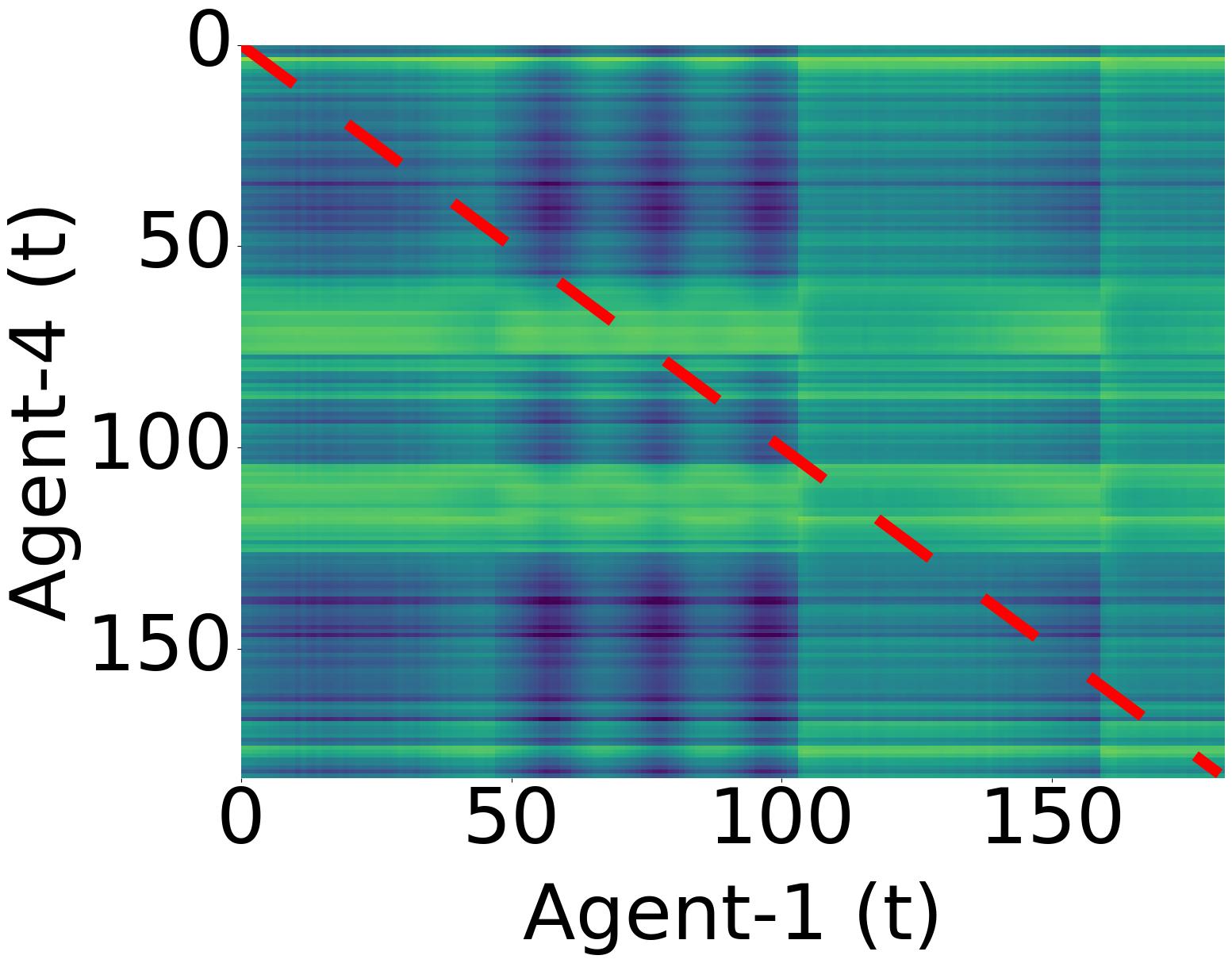}
    \end{subfigure}
    \begin{subfigure}[b]{0.19\linewidth}
        \centering
        \includegraphics[width=\linewidth]{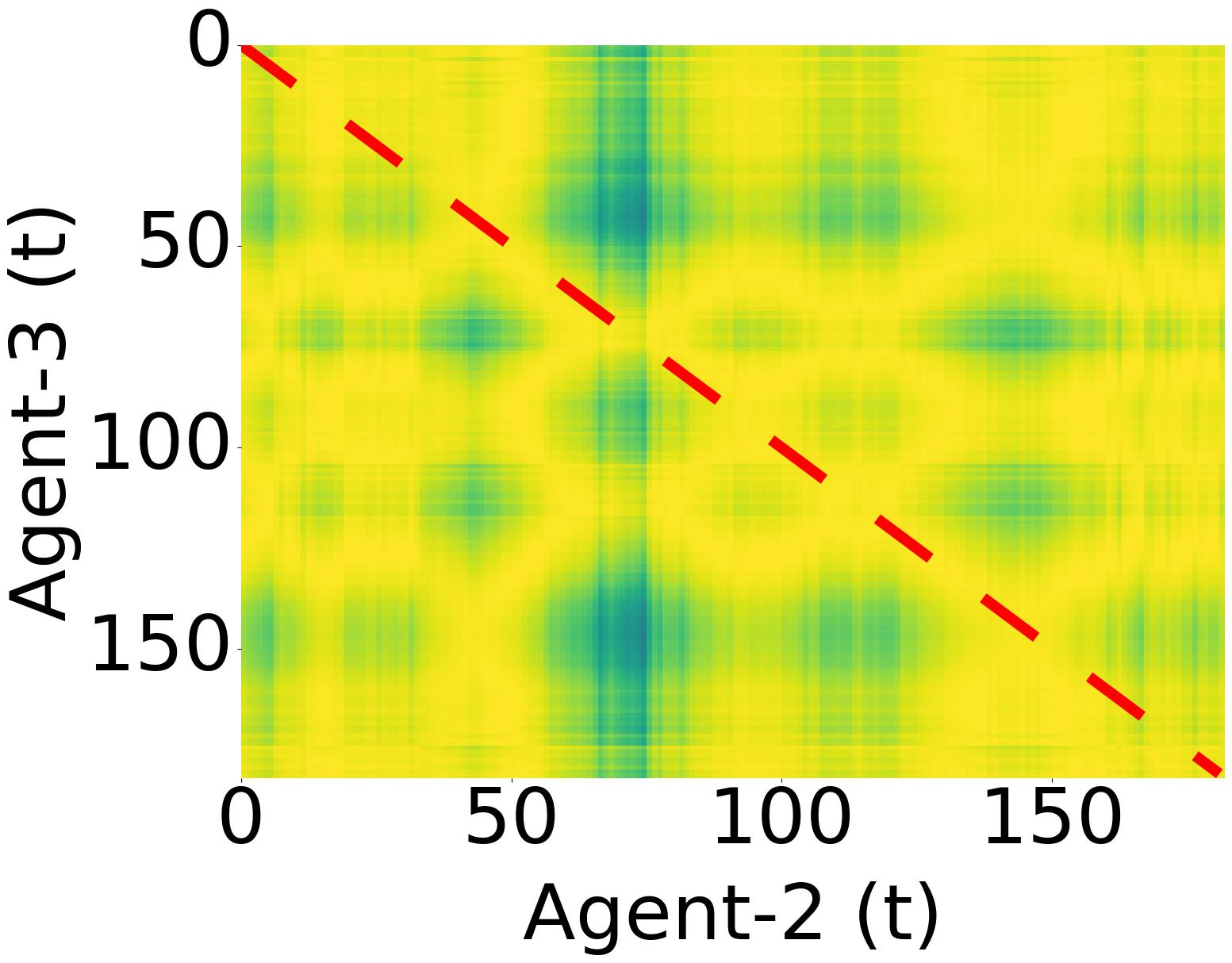}
    \end{subfigure}
        \begin{subfigure}[b]{0.19\linewidth}
        \includegraphics[width=\linewidth]{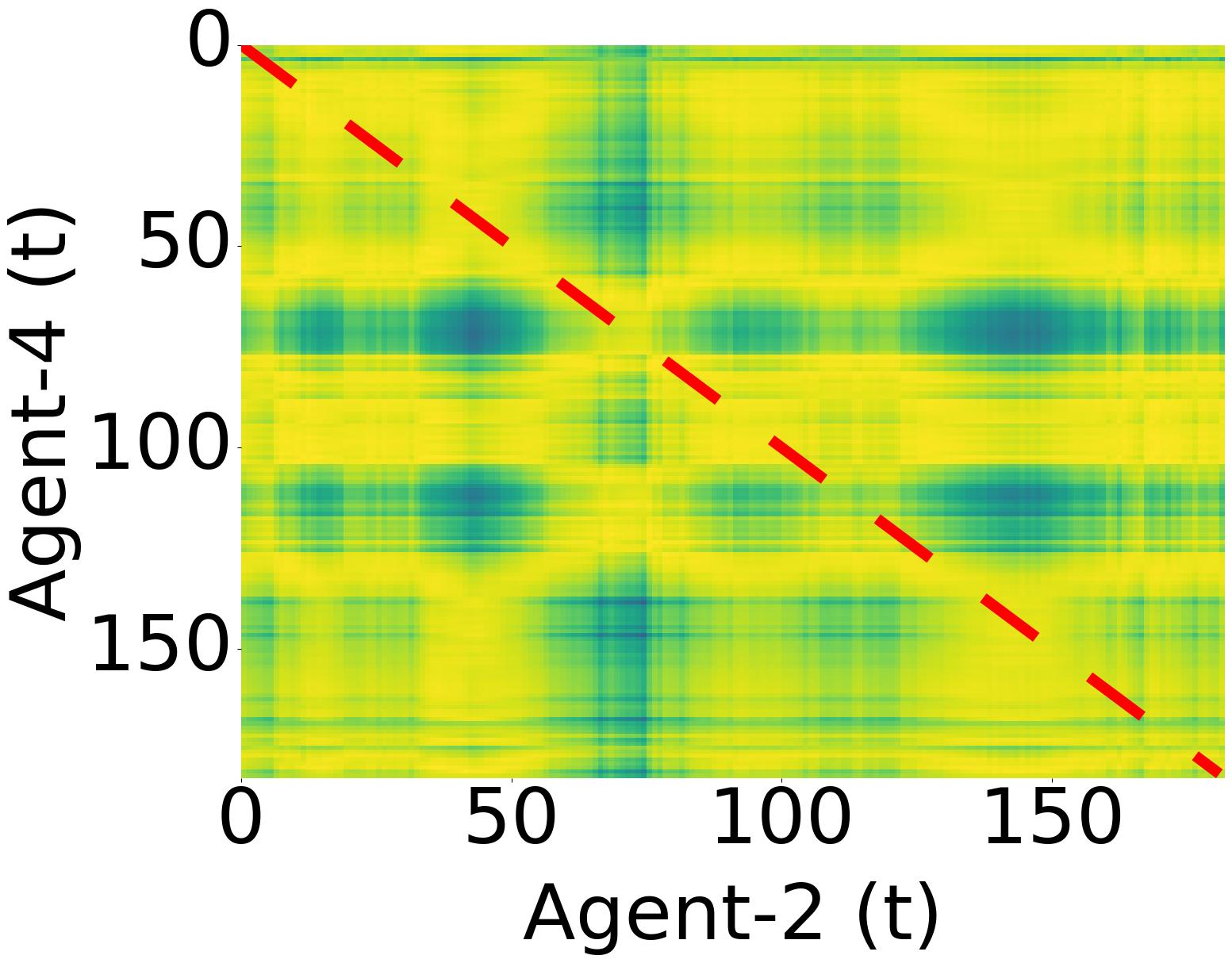}
    \end{subfigure}
    \begin{subfigure}[b]{0.19\linewidth}
        \includegraphics[width=\linewidth]{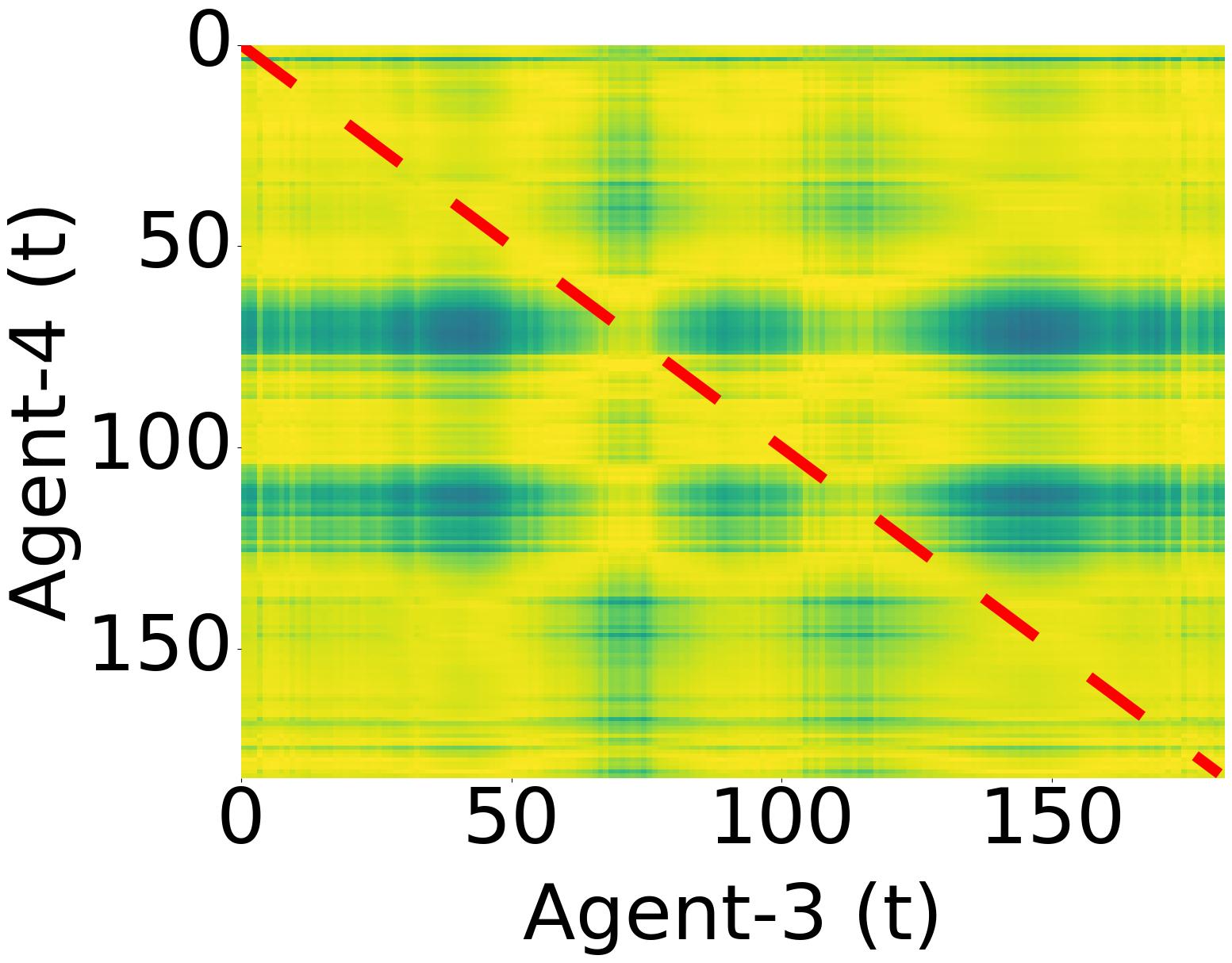}
    \end{subfigure}
    \caption{Embedding similarity matrices in the complex context after \num{30} epochs, where yellow indicates the highest similarity, gradually decreasing with darker shades.}
    \label{fig:experiment-2-similarity}
\end{figure*}
\subsection{Robustness of Learning}
In the second experiment, we introduced two \textit{misleading agents}, $Agent_{M0}$ and $Agent_{M1}$, into the simple context (\ref{p:simple-context}) with $Agent_0$ and $Agent_1$, whose data are randomly generated from a normal distribution with mean \num{0} and standard deviation \num{1} to intentionally introduce uncorrelated information and disrupt the alignment process. All four agents are trained for five epochs, achieving a maximum accuracy of \num{0.9935} at the last epoch. 

As shown in Fig. \ref{fig:convergence-with-noise}, the similarity matrix of $Agent_0$ and $Agent_1$ retains the patterns observed in Fig.~\ref{fig:experiment-0-similarity}, demonstrating their ability to preserve local embedding alignment despite disturbances. In contrast, the similarity matrices of $Agent_{M0}$ and $Agent_{M1}$ show disorganized patterns without a dominant main diagonal, indicating their inability to generate consistent representations due to random data.
This experiment thus confirms that \OLshort effectively maintains representation alignment in agents with correlated data while isolating the impact of unstructured, semantically irrelevant contributions.

\subsection{Emerging of Sub-Contexts}
In the final experiment, we trained the five agents in the complex context (\ref{p:complex-context}), i.e., $Agent_0$--$Agent_4$, for \num{30} epochs. As shown in Fig.~\ref{fig:experiment-2-similarity}, the embedding similarity matrices indicate that $Agent_0$ and $Agent_1$ maintain alignment despite the applied offset. Similarly, $Agent_2$ and $Agent_3$, despite their negative correlation, converge to a common representation. Moreover, $Agent_4$, despite only partial correlation, successfully aligns with $Agent_2$ and $Agent_3$ by leveraging the information contained in its feature pair.
\begin{figure}[!t]
    \begin{subfigure}[b]{0.325\linewidth}
  \includegraphics[width=\linewidth]{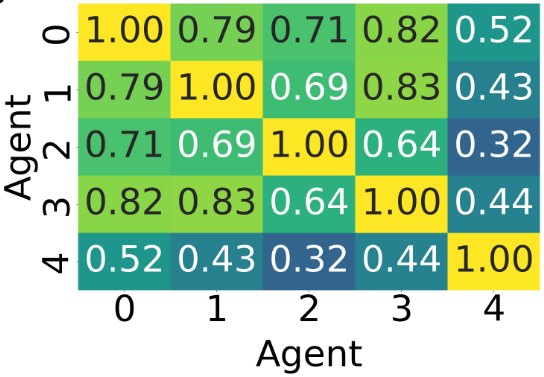}
        \caption{Epoch 2}
        \label{fig:clustering-evolution-a}
    \end{subfigure}
    \begin{subfigure}[b]{0.325\linewidth}
        \includegraphics[width=\linewidth]{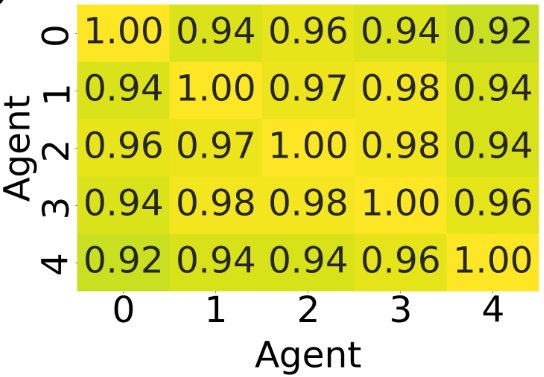}
        \caption{Epoch 16}
        \label{fig:clustering-evolution-b}
    \end{subfigure}
    \begin{subfigure}[b]{0.325\linewidth}
        \centering
        \includegraphics[width=\linewidth]{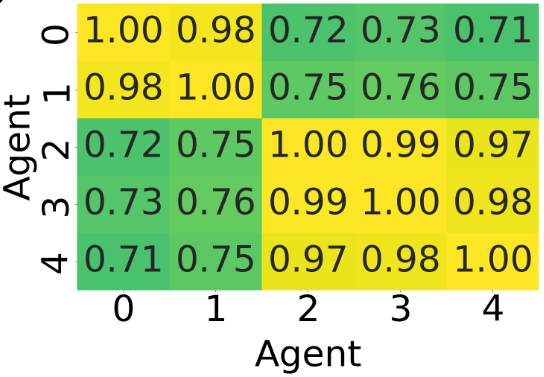}
        \caption{Epoch 30}
        \label{fig:clustering-evolution-c}
    \end{subfigure}
    
    \caption{Average embedding similarity at epochs \num{2} (a), \num{16} (b), and \num{30} (c), showing the formation of distinct sub-contexts for correlated groups ($Agent_0$,$Agent_1$) and ($Agent_2$--$Agent_4$).}
    \label{fig:clustering-evolution}
\end{figure}
\begin{figure}[!t]
    \centering
    \begin{subfigure}[b]{\linewidth}
        \centering
        \caption{Training loss}
        \includegraphics[width=0.652\linewidth]{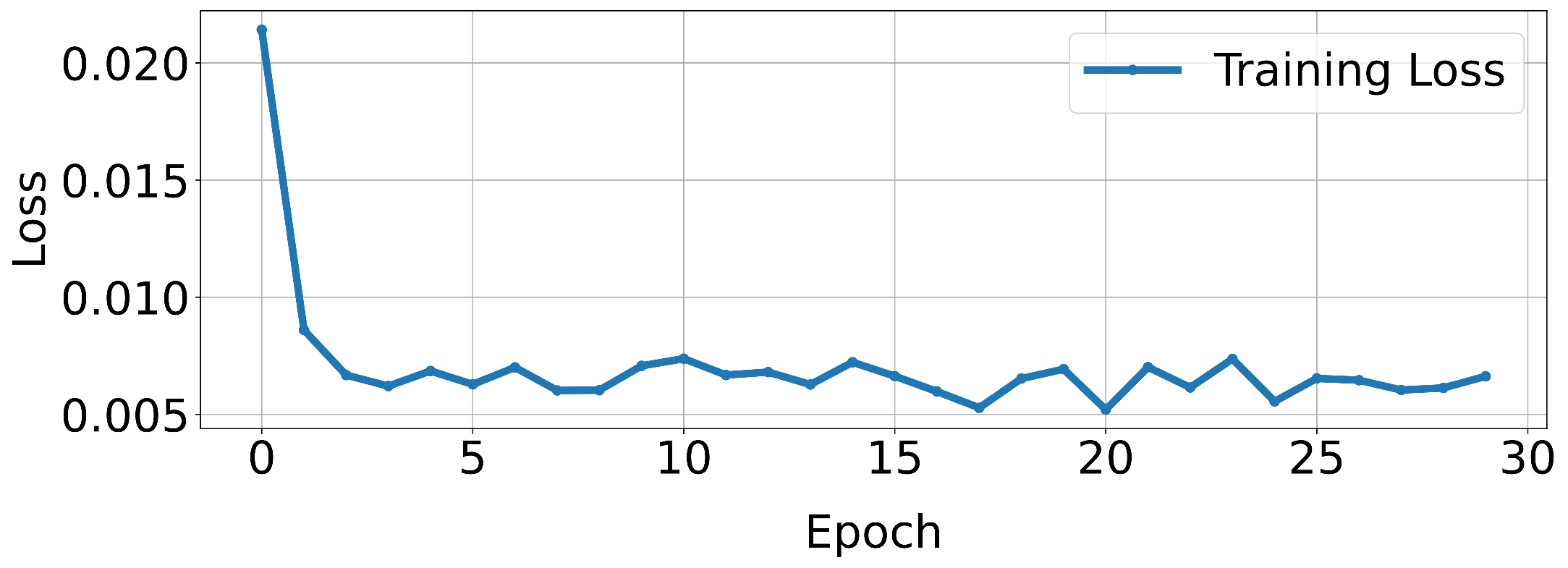}
    \end{subfigure}
    \begin{subfigure}[b]{\linewidth}
        \centering
        \caption{Accuracy evolution}
        \includegraphics[width=0.652\linewidth]{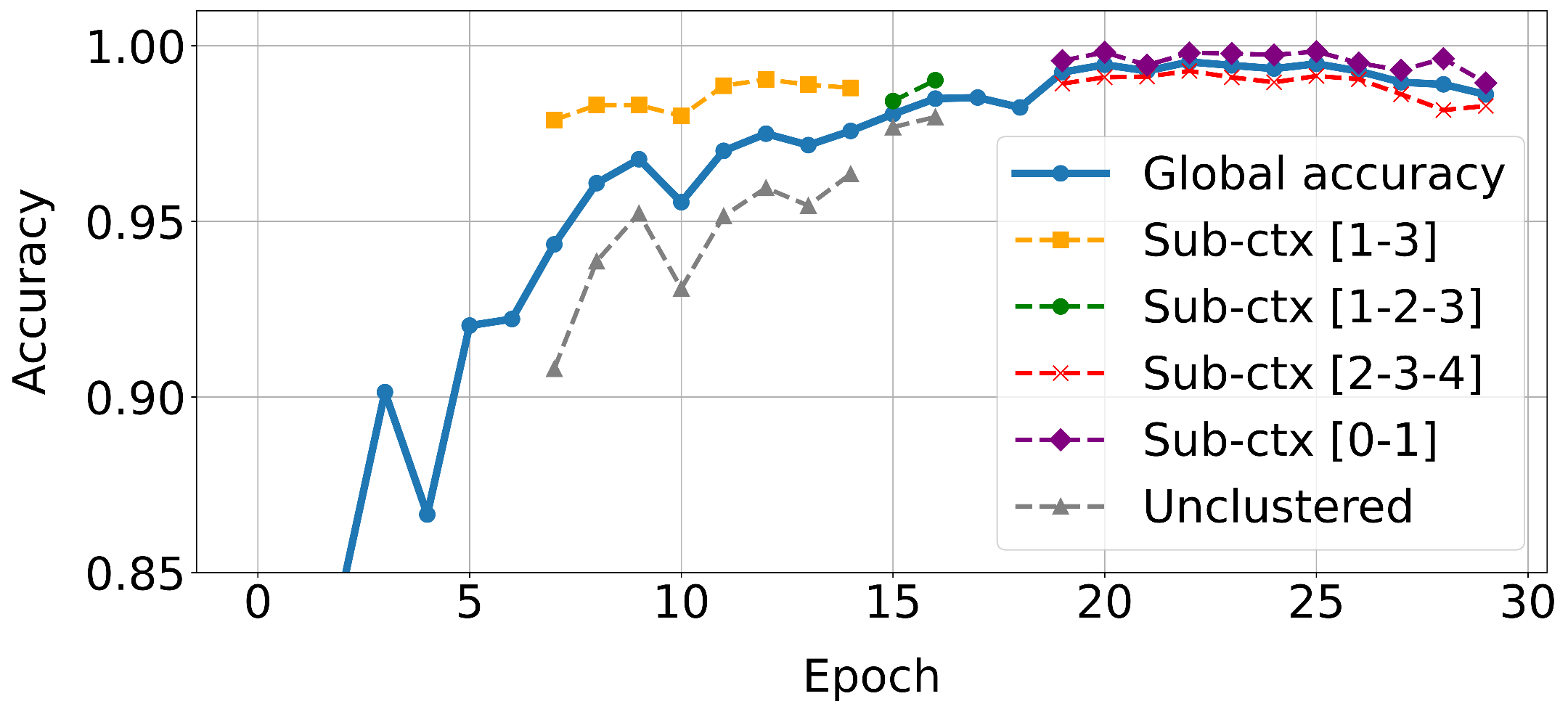}
    \end{subfigure}
    \caption{Training loss (a) and accuracy evolution (b) over \num{30} epochs, highlighting convergence and sub-contexts emergence.}
    \label{fig:experiment-2-loss-accuracy}
\end{figure}

Fig.~\ref{fig:clustering-evolution} shows the evolution of clusters. Initially, all agents exhibit poorly defined representations, gradually converging toward a common representation driven by the embedding alignment loss function (Eq.~\ref{eq:loss-alignment}). Once local embeddings reach sufficient similarity, the local information preservation loss function (Eq.~\ref{eq:loss-preservation}) ensures information retention, leading to the emergence of distinct sub-contexts.

As a final investigation, we analyzed the training loss and accuracy trends over the epochs, as shown in Fig.~\ref{fig:experiment-2-loss-accuracy}. The rapid decline in training loss during the early epochs indicates effective initial learning, followed by convergence. In addition, global accuracy rises sharply, peaking at \num{0.9954} at epoch \num{23}, demonstrating strong generalization across the datasets.
Beyond overall performance, Fig.~\ref{fig:experiment-2-loss-accuracy} also reveals insights into sub-context formation. While transient, unexpected sub-contexts emerge between $Agent_1$ and $Agent_2$ as well as $Agent_1$ and $Agent_3$ at intermediate stages, the correct sub-contexts ($Agent_0$--$Agent_1$ and $Agent_2$--$Agent_4$) eventually stabilize, each following distinct learning trajectories, highlighting \OLshort's ability to dynamically structure representations based on underlying data relationships.

%% file: tex/7-sota.tex
\section{How \OLshort Relates to Existing Paradigms}
\label{sec:sota}
This section reviews relevant paradigms and distinguishes \OLshort's novelties. 
\paragraph{Representation Learning} fitting with representation learning algorithms~\cite{6472238}, \OLshort leverages latent representations to extract information. Like many unsupervised representation learning methods \cite{uelwer2023surveyselfsupervisedrepresentationlearning}, \OLshort employs a distributed, self-supervised process. However, instead of creating targets to represent data, \OLshort focuses on uncovering contextual structures, distinguishing them from conventional methods.
\paragraph{Federated Learning} adopting a worker-master architecture used in Federated Learning (FL), \OLshort differs significantly in objectives and methodology. In FL, workers train local models independently, and a central server aggregates model parameters to build a globally generalized model \cite{mcmahan2023communicationefficientlearningdeepnetworks, Shanmugam_Tillu_Tomar_2023}. In contrast, \OLshort does not share model parameters; instead, it exchanges local model outputs, emphasizing contextual representation rather than global model aggregation.
\paragraph{Knowledge Distillation} focusing on aligning model outputs across distributed entities, \OLshort shares similarities with Knowledge Distillation (KD)~\cite{hu2023teacherstudentarchitectureknowledgedistillation} but differs fundamentally in its approach. While KD relies on a teacher model to guide student models toward supervised convergence, \OLshort operates as a fully self-supervised process. The global context in \OLshort emerges naturally from correlations among distributed data, without an explicit supervisor. Although unsupervised KD variants exist \cite{10433659}, their objectives differ: KD focuses on transferring task-specific knowledge across models, whereas \OLshort leverages distributed learning to extract contextual representations.
\paragraph{Contrastive Learning} aiming to align representations through implicit relationships, \OLshort exhibits similarities with Contrastive Learning (CL) \cite{10.1145/3637528.3671454}. However, while CL relies on multiple views or negative pairs to enforce similarity among related representations and dissimilarity among unrelated ones \cite{xu2022negativesamplingcontrastiverepresentation}, \OLshort removes this requirement. Instead, it aligns local representations by leveraging latent correlations, enhancing its efficiency and suitability for distributed environments.
\paragraph{Distributed Clustering} sharing the goal of identifying similarities in datasets, \OLshort aligns with Distributed Clustering methods like K-Means and Density-Based Spatial Clustering of Applications with Noise (DBSCAN)~\cite{9038535}. However, while distributed clustering simply partitions data across nodes, \OLshort integrates clustering as an inherent outcome of the learning process. During training, sub-contexts emerge dynamically among strongly related agents, forming semantic clusters without requiring explicit supervision.
\paragraph{Multi-Agent Learning} building on the concept of independent yet coordinated agents, \OLshort aligns with Multi-Agent Reinforcement Learning (MARL) \cite{NING202473}. However, while MARL optimizes agent policies to maximize rewards or achieve a collective goal, \OLshort focuses exclusively on aligning representations to construct latent shared knowledge.
\paragraph{Knowledge Graphs} aligning in representing relationships among distributed data \cite{10.1145/3618295}, \OLshort differs from Knowledge Graphs (KG) in its approach. While KGs explicitly structure relationships in a supervised manner, requiring prior knowledge of data connections, \OLshort allows them to emerge autonomously during training. This enables the capture of latent correlations without predefined or explicit definitions.
\paragraph{Autoencoder} aligning with Autoencoders \cite{Berahmand2024} in generating latent representations, \OLshort differs in approach by focusing on contextual representation rather than data reconstruction. While autoencoders compress and reconstruct data, \OLshort builds a shared contextual representation without recovering the original input. However, like autoencoders, it can incorporate a reconstruction loss by balancing global alignment and local information preservation (Eq. \ref{eq:loss}). Variational Autoencoders (VAEs) \cite{10.1145/3663364}, though unsupervised, model probabilistic latent spaces for data compression, reconstruction, and sample generation. In contrast, \OLshort forgoes probabilistic modeling and generation, focusing solely on the semantic alignment of local representations.

%% file: tex/8-discussion.tex
\section{Discussion}
\label{sec:discussion}
\OLshort introduces a novel paradigm for self-supervised distributed learning, enabling agents to extract latent contextual knowledge from decentralized data. While the approach demonstrates strong representation alignment and sub-context formation, several technical challenges remain in scalability, efficiency, and real-world deployment. This section discusses three key limitations and future research directions.
\paragraph{Integration with attention mechanisms}
osmosis and self-attention mechanism~\cite{NIPS2017_3f5ee243} are complementary, as the latter leverages existing contextual information to refine and enrich data representations, while osmosis operates in the opposite direction, extracting and synthesizing contextual knowledge from distributed data. Exploring attention-driven weighting within \OLshort{}’s embedding alignment could refine representations and improve adaptability.
\paragraph{Processed vs. raw data for representation learning} \OLshort currently operates on raw data, which may limit efficiency in structured domains. Investigating preprocessed feature embeddings~\cite{wu2024deepfeatureembeddingtabular}, task-specific transformations~\cite{ziko2023taskadaptivefeaturetransformation}, and feature importance weighting~\cite{Chan2024} could enhance interpretability and reduce training complexity.
\paragraph{Real-world deployment and scalability}
deploying \OLshort across distributed environments requires efficient synchronization mechanisms and privacy-preserving embedding aggregation. Further optimizations in adaptive resource allocation will also be essential for large-scale adoption. Therefore, future work should explore low-rank projections, hierarchical aggregation, and asynchronous training to improve efficiency while maintaining alignment stability.

%% file: tex/9-conclusions.tex
\section{Conclusions}
\label{sec:conclusion}
This paper introduced \OSL (\OLshort), a self-supervised distributed learning paradigm that extracts latent contextual knowledge from decentralized data. By aligning local embeddings into a shared representation, \OLshort converges effectively handling data heterogeneity without centralizing raw information. The experimental results demonstrated that \OLshort consistently generated robust representations, even in the presence of noise or partial correlations, and dynamically identified sub-contexts through dynamic clustering, achieving an accuracy of \num{0.99} within each group. 
As an emerging paradigm, \OLshort opens new research perspectives, e.g., with a future focus on enhancing \OLshort features and extending its applicability to large-scale, real-world systems. 